%% file: main.tex
\documentclass[10pt,twocolumn,letterpaper]{article}

\usepackage[pagenumbers]{cvpr} 


\usepackage{colortbl}

\definecolor{lightblue}{rgb}{0.85,0.92,1.0} 

\newcommand{\DatasetName}{BiAudio}
\newcommand{\MethodName}{ViSAudio}

\definecolor{cvprblue}{rgb}{0.21,0.49,0.74}
\usepackage[pagebackref,breaklinks,colorlinks,allcolors=cvprblue]{hyperref}
\usepackage{multirow}
\usepackage{amssymb}  
\usepackage{marvosym}
\usepackage{algorithm}
\usepackage{algpseudocode}

\def\paperID{3743} 
\def\confName{CVPR}
\def\confYear{2026}

\title{\MethodName{}: End-to-End Video-Driven Binaural Spatial Audio Generation}

\author{Mengchen Zhang$^{1,2}$, Qi Chen$^{3,4}$, Tong Wu$^{5}$\textsuperscript{\Letter}, Zihan Liu$^{6,2}$, Dahua Lin$^{2,7}$\textsuperscript{\Letter}\\
\normalsize $^{1}$Zhejiang University, 
$^{2}$Shanghai Artificial Intelligence Laboratory,  \\
\normalsize $^{3}$Shanghai Jiao Tong University,
$^{4}$Shanghai Innovation Institute,
$^{5}$Stanford University,\\
\normalsize $^{6}$Beihang University,
$^{7}$The Chinese University of Hong Kong\\
\tt\small zhangmengchen@zju.edu.cn,
cq1073554383@sjtu.edu.cn,
wutong16@stanford.edu,\\ \tt\small
liuzihan@buaa.edu.cn,
dhlin@ie.cuhk.edu.hk,
 }

\begin{document}
\input{Figures/teaser}
\input{sec/0_abstract}    
\input{sec/1_intro}
\input{sec/2_related_work}
\input{sec/3_dataset}

\input{sec/4_method}

\input{sec/5_experiment}

\input{sec/6_conclusion}



\appendix


\section*{Appendix}

\setcounter{table}{0}
\setcounter{figure}{0}
\setcounter{footnote}{0}
\renewcommand{\thesection}{\Alph{section}}
\renewcommand\thefigure{R\arabic{figure}}
\renewcommand\thetable{S\arabic{table}}

\maketitle
\input{sec/supplementary}

{
    \small
    \bibliographystyle{ieeenat_fullname}
    \bibliography{main}
}
\end{document}

%% file: Figures/teaser.tex
\twocolumn[{
\renewcommand\twocolumn[1][]{#1}
\maketitle
\begin{center}
    \vspace{-10pt}
    \centering
    \includegraphics[width=\linewidth]{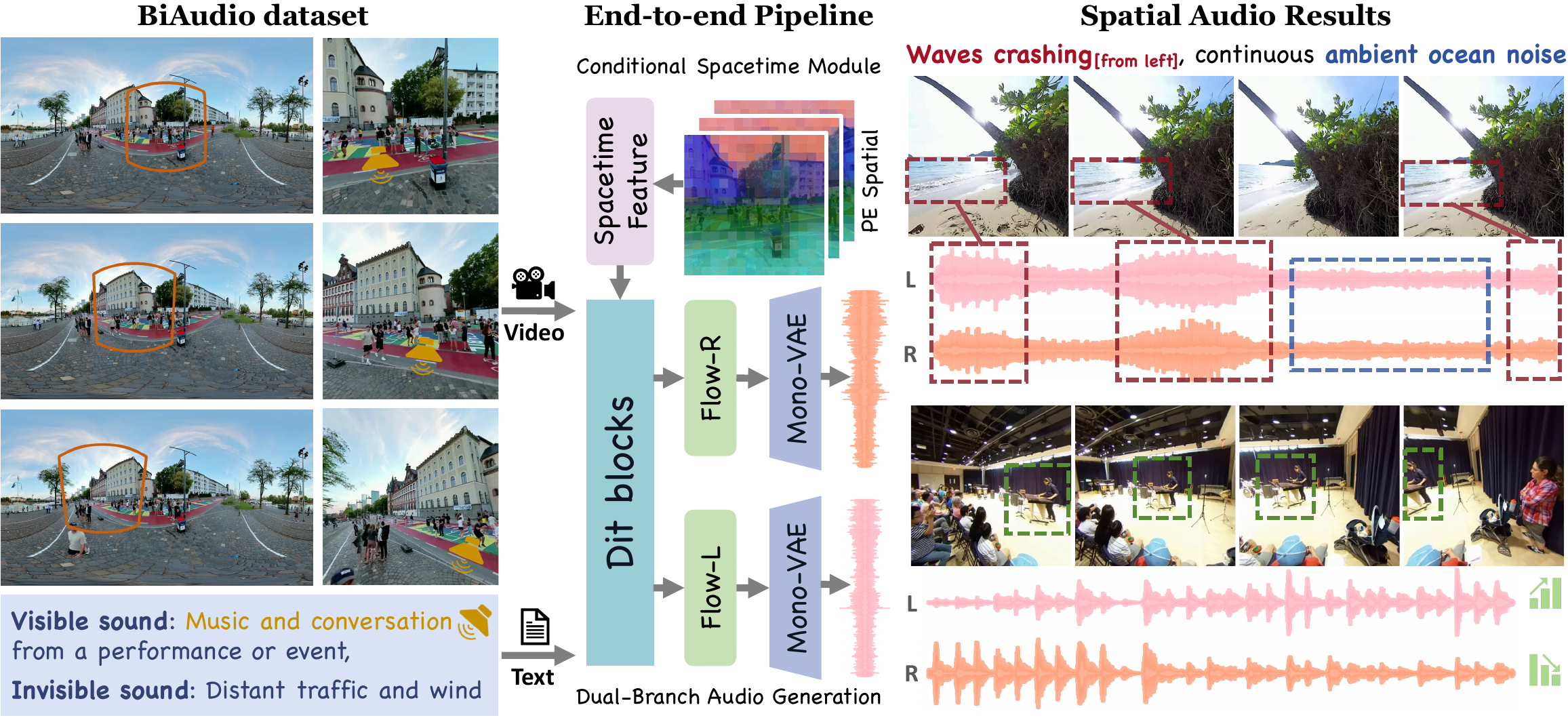}
    \setlength{\abovecaptionskip}{0mm}
    \vspace{-8pt}
    \captionof{figure}
    {\textbf{Overview.} 
    \textbf{Left}: \DatasetName{} dataset converts 360$^\circ$ videos and FOA audio into perspective video and binaural audio pairs, employing diverse camera rotations to enhance spatial cues.
    \textbf{Middle}: Our end-to-end pipeline employs conditional flow matching with a dual-branch generation architecture, integrated with a conditional spacetime module to generate spatially immersive binaural audio from multimodal inputs. \textbf{Right}: Example results generated by \MethodName{}. 
    As shown above, our model faithfully generates the visible sound of \textcolor[rgb]{0.6471,0.0667,0.1451}{\textbf{waves crashing}}, highlighted with red boxes in both the video frames and the audio waveform, with the left channel louder since the sound event occurs on the left. It also captures subtle environmental sounds like \textcolor[rgb]{0.1804,0.3294,0.6314}{\textbf{ocean noise}}, highlighted with blue boxes, demonstrating its ability to reproduce fine-grained background acoustics. 
    As shown below, as the camera rotates right, the \textcolor[rgb]{0.3451,0.5569,0.1922}{\textbf{marimba sound}} moves left, increasing left-channel amplitude while decreasing the right, demonstrating dynamic adaptation to viewpoint changes.}
    \vspace{3pt}
	\label{fig:teaser}
\end{center}
}]

%% file: sec/0_abstract.tex
\begin{abstract}
Despite progress in video-to-audio generation, the field focuses predominantly on mono output, lacking spatial immersion. Existing binaural approaches remain constrained by a two-stage pipeline that first generates mono audio and then performs spatialization, often resulting in error accumulation and spatio-temporal inconsistencies.
To address this limitation, we introduce the task of \textbf{end-to-end binaural spatial audio generation} directly from silent video. To support this task, we present the \textbf{\DatasetName{}} dataset, comprising approximately 97K video-binaural audio pairs spanning diverse real-world scenes and camera rotation trajectories, constructed through a semi-automated pipeline.
Furthermore, we propose \textbf{\MethodName{}}, an end-to-end framework that employs conditional flow matching with a dual-branch audio generation architecture, where two dedicated branches model the audio latent flows. Integrated with a conditional spacetime module, it balances consistency between channels while preserving distinctive spatial characteristics, ensuring precise spatio-temporal alignment between audio and the input video.
Comprehensive experiments demonstrate that \MethodName{} outperforms existing state-of-the-art methods across both objective metrics and subjective evaluations, generating high-quality binaural audio with spatial immersion that adapts effectively to viewpoint changes, sound-source motion, and diverse acoustic environments.
Project website: \href{https://kszpxxzmc.github.io/ViSAudio-project}{\textit{https://kszpxxzmc.github.io/ViSAudio-project}}.
\end{abstract}

%% file: sec/1_intro.tex
\section{Introduction}

With the rapid growth of virtual and augmented reality~\cite{virtual, Augmented}, the demand for realistic, immersive audio-visual experiences has surged. To achieve immersion, sound must not only be synchronized with visual content but also convey spatial awareness. 
Binaural spatial audio creates a highly realistic spatial listening experience by simulating a two-dimensional soundscape. However, traditional binaural audio production requires specialized equipment and expertise~\cite{Ambisonics}. Therefore, automatically generating spatial binaural audio from silent video offers substantial practical value.

Driven by advancements in generative AI, recent research~\cite{Diff-Foley, LoVA, Seeing-and-Hearing, VTA-LDM, Kling-Foley, FoleyCrafter, MultiFoley, TA-V2A} has made significant strides in generating mono audio from silent video in an end-to-end manner. However, binaural audio generation remains constrained by a two-stage process: mono audio is first generated using a pretrained video-to-mono-audio generator and then transformed into binaural spatial audio through a separate spatialization process. Some methods~\cite{Sonic4D, See-2-sound, FoleySpace} localize and track sound sources in input video to synthesize plausible spatial audio, while others~\cite{SpatialAudioGen, Geometry, Mono2Binaural, Sep-stereo, PseudoBinaural} employ UNet-like~\cite{Unet} architectures to predict binaural channels directly from mono audio.
However, these approaches rely heavily on pre-synthesized mono audio, making their performance inherently constrained by the quality of the first-stage output. Meanwhile, the subsequent spatialization often considers only visible sound sources, ignoring off-screen sounds and environmental noise. This two-stage paradigm is therefore prone to error accumulation, leading to misalignment with the input video and inconsistencies in spatial.
Recent methods~\cite{OmniAudio, ViSAGe} have established an end-to-end paradigm for spatial audio generation, enabling direct synthesis of First-Order Ambisonics (FOA) from video. However, these approaches depend on 360$^\circ$ video inputs or rely on extra parameters such as camera orientation. 
In contrast, end-to-end binaural audio generation aims to extract spatial cues directly from perspective video, making it broadly applicable. Nevertheless, this direction remains largely unexplored.

In this work, we propose to generate \textbf{binaural} spatial audio from silent video in an \textbf{end-to-end} manner. 
This is a challenging task due to data sparsity: Existing real-world video-binaural datasets~\cite{OAP, Fair-Play,MUSIC1, MUSIC2} are small in scale, lack diversity, and often focus on narrow environments like street scenes or music. Others rely on synthetic audio and video~\cite{SimBinaural}, limiting their real-world applicability. Moreover, previous datasets are often constrained by fixed camera perspectives, with very few containing camera motion. To address these limitations, we introduce \textbf{\DatasetName{}}, a large-scale, open-domain dataset featuring diverse real-world sound environments. It consists of approximately 97,000 pairs of binaural spatial audio and video clips, each lasting 8 seconds and accompanied by descriptive captions, totaling 215 hours. We developed a semi-automated pipeline for dataset construction, during which we diversified camera rotation trajectories to ensure that the generated audio is not restricted by fixed viewpoints.

Furthermore, we propose \textbf{\MethodName{}}, the first end-to-end framework designed to generate binaural spatial audio based on silent video and optional text conditions. 
\MethodName{} employs conditional flow-matching to jointly model the left and right audio channels through a \textit{Dual-Branch Audio Generation} design, ensuring channel consistency while maintaining distinct spatial characteristics. To further improve spatio-temporal fidelity, we introduce a \textit{Conditional Spacetime Module} that extracts synchronization and spatial features from the video and injects them into the dual-channel generation process. The architecture enables the generation of high-quality, immersive audio that remains spatio-temporally aligned with the input video, outperforming existing state-of-the-art approaches.

Our contributions can be summarized as follows:

\begin{itemize}
\item We curate \DatasetName{}, a large-scale, open-domain dataset of video–binaural audio pairs with diverse camera motions, along with a semi-automated construction pipeline.

\item We propose \MethodName{}, a novel framework that integrates a conditional spacetime module into dual-branch audio generation, achieving channel coherence and spatial distinctiveness with precise audio–visual spatial consistency. We achieve end-to-end binaural audio generation from silent video, bypassing the limitations of conventional two-stage approaches.

\item Extensive experiments demonstrate that \MethodName{} outperforms state-of-the-art methods on both objective metrics and subjective evaluations, generating high-quality binaural audio with strong spatial immersion and adapting effectively to viewpoint changes, sound-source motion, and diverse acoustic environments.

\end{itemize}

\label{sec:intro}

%% file: sec/2_related_work.tex
\input{Tables/dataset}

\section{Related Work}
\label{sec:related_work}
\noindent\textbf{Video-to-Audio Generation.}
Early research~\cite{SpecVQGAN, Foleygen, V-AURA, MaskVAT, CODI2} primarily focuses on auto-regressive models. SpecVQGAN~\cite{SpecVQGAN} pioneers open-domain video-to-audio generation using a VQGAN-based~\cite{VQGAN} Mel-spectrogram codebook. 
Diffusion models~\cite{Diffusion} significantly advance V2A generation~\cite{Diff-Foley, LoVA, Seeing-and-Hearing, VTA-LDM, Kling-Foley, FoleyCrafter, MultiFoley, TA-V2A}. Diff-Foley~\cite{Diff-Foley} combines contrastive audio–visual pretraining with latent diffusion to model spectrogram representations in latent space.
Recent works~\cite{Frieren, MovieGen, MMAudio, ThinkSound} adopt flow-matching-based~\cite{Flow-Matching, FM2} generative models. 
Frieren~\cite{Frieren} applies rectified flow matching~\cite{RFM} with reflow and one-step distillation for efficiency. MMAudio~\cite{MMAudio} employs a flow-matching framework conditioned on multimodal inputs.
Emerging multimodal frameworks~\cite{See-Hear-Read, Kling-Foley, MovieGen, FoleyCrafter, CODI2, MMAudio, ThinkSound, MultiFoley, TA-V2A} further advance the field by jointly modeling audio, visual, and textual modalities.
Our method, \MethodName{}, goes further by leveraging spatial cues to generate binaural audio, employing a flow-matching framework conditioned on multimodal inputs.

\noindent\textbf{Visual-based Spatial Audio Generation.}
Producing spatial audio typically requires specialized equipment (\eg, microphone arrays), motivating research into generating it automatically from visual information.
Early approaches~\cite{Sonic4D, See-2-sound, FoleySpace, Ccstereo, SpatialAudioGen, Geometry, Mono2Binaural, Sep-stereo, PseudoBinaural} typically follow a two-stage pipeline for visual-based spatial audio generation: mono audio is first generated and then transformed into spatial audio using visual cues. Some methods~\cite{Sonic4D, See-2-sound, FoleySpace} localize and track sound sources from input videos to synthesize plausible spatial audio, while others~\cite{Ccstereo, SpatialAudioGen, Geometry, Mono2Binaural, Sep-stereo, PseudoBinaural} adopt UNet–like~\cite{Unet} architectures to predict binaural channels from mono audio directly. However, these methods rely heavily on pre-existing mono audio inputs, introducing additional challenges such as misalignment with visuals and limited spatial consistency.
Recently, end-to-end methods~\cite{OmniAudio, ViSAGe} have emerged, directly generating spatial audio by leveraging both spatial and semantic cues from videos. OmniAudio~\cite{OmniAudio} generates first-order ambisonics (FOA) audio from 360° panoramic videos, while ViSAGe~\cite{ViSAGe} generates FOA audio from field-of-view (FoV) videos conditioned on corresponding camera directions. However, end-to-end binaural audio generation remains unexplored. We propose the first end-to-end framework that directly infers spatially consistent binaural audio from visual content.

%% file: Tables/dataset.tex
\begin{table*}
\centering
\small
\vspace{-8pt}
\caption{\textbf{\DatasetName{} Dataset.} Comparison between \DatasetName{} and existing binaural audio-video datasets. FoV and 360$^\circ$ denote Field-of-View and panoramic videos, respectively, while FOA stands for First-order Ambisonics. \DatasetName{} is currently the largest video-binaural audio dataset, featuring open-domain sounds from diverse real-world environments and varied camera rotation trajectories, enabling audio generation beyond fixed viewpoints. Rich captions allow modeling of subtle environmental sounds, producing more realistic audio.}

\vspace{-7pt}
\begin{tabular}{@{}l|cc|cccc|cc|c@{}}
\toprule
\multirow{2}{*}{\textbf{Dataset}} & \multicolumn{2}{c|}{\textbf{Statistic}} & \multicolumn{4}{c|}{\textbf{Data Type}} & \multicolumn{2}{c|}{\textbf{Caption}} &  \textbf{Camera} \\
 & \textbf{\#Clips} & \textbf{Duration} & \textbf{Video Type} & \textbf{Open Domain} & \textbf{Real-world} & & \textbf{Visible} & \textbf{Invisible} & \textbf{Viewpoint} \\
\midrule
OAP~\cite{OAP}             & 64k     & 26h      & 360$^\circ$        & Street   & \checkmark & & $\times$ & $\times$ & Fixed \\
Fair-Play~\cite{Fair-Play} & 1.9k    & 5.2h     & FoV                & Music    & \checkmark & & $\times$ & $\times$ & Fixed  \\
SimBinaural~\cite{SimBinaural}  & 22k     & 116.1h   & FoV                & Music    & $\times$   & & $\times$ & $\times$ & Fixed  \\
MUSIC-21~\cite{MUSIC1,MUSIC2}   & 1.7k    & 81h      & FoV                & Music    & \checkmark & & \checkmark & $\times$ & Fixed \\
YouTube-Binaural~\cite{SimBinaural} & 0.4k & 27h    & FoV                & \checkmark & \checkmark & & $\times$ & $\times$ & Fixed \\
\textbf{\DatasetName{} (Ours)}     & \textbf{97k} & \textbf{215h} & \textbf{FoV} & \checkmark & \checkmark & & \checkmark & \checkmark & \textbf{Moving} \\
\bottomrule
\end{tabular}
\label{tab:dataset}
\end{table*}

%% file: sec/3_dataset.tex
\section{\DatasetName{} Dataset}
\label{sec:dataset}

We introduce \textbf{\DatasetName{}}, a large-scale dataset for video-to-spatial audio generation.
It consists of about 97,000 pairs of binaural spatial audio and perspective video clips, each lasting 8 seconds and accompanied by descriptive captions, for a total of 215 hours.
To ensure perceptually distinct spatial cues, we generate diverse camera rotation trajectories and discard samples with minimal left-right channel differences. The data construction process is detailed in \cref{sec:Construction}, and the dataset statistics are presented in \cref{sec:Statistics}.

\subsection{Dataset Construction}
\label{sec:Construction}

\noindent\textbf{Data Creation.} 
We construct our dataset by sampling paired first-order ambisonics (FOA) audio and 360° video from the Sphere360 dataset~\cite{OmniAudio}, which covers a diverse range of real-world, open-domain acoustic environments. 
To better align with human perception, the 360° videos are projected into 90° perspective views along pre-defined camera rotation trajectories, while the corresponding FOA audio is rendered into binaural signals via Head-Related Impulse Response (HRIR) convolution. 

However, in real-world perception, humans can hear off-screen sounds even when observing a limited perspective view. Accordingly, real-world binaural audio contains abundant off-screen sounds as well as environmental noises that cannot be localized. 
As illustrated on the left of ~\cref{fig:teaser}, humans can localize the sound of \textit{music and conversation}, which is visible in the video and marked with yellow speaker icons, whereas off-screen sounds such as \textit{distant traffic} and environmental noises like \textit{wind} cannot be precisely localized.
Using such raw audio-visual data indiscriminately may introduce semantic noise and weaken audio-visual alignment.
To address this, we design a two-stage pipeline to finely annotate visible and invisible sound sources. Specifically, Qwen2.5-Omni~\cite{Qwen2.5-Omni} produces detailed textual descriptions that capture both visible sounds and background sounds, including off-screen sources and environmental noise. These descriptions are then distilled by Qwen3-Instruct-2507~\cite{qwen3} into concise captions following the format: \textit{``Visible sound:\;\textless audible sound\textgreater, Invisible sound:\;\textless background sound\textgreater''}. A sample is shown in ~\cref{fig:teaser}. For details, please refer to Appendix~\cref{sec:Construction_sup}

\noindent\textbf{Spatial Cue Enhancement.}
\label{sec:Enhancement}
To strengthen perceptually distinct spatial cues and overcome the constraints of fixed viewpoints, we diversify camera rotation trajectories. Each trajectory starts from a random pitch, yaw, and roll orientation, followed by random drifts to mimic natural movements. Additionally, we ensure that the main visual cues corresponding to the audio stay within the field of view for a certain duration. Specifically, we compute the direction of maximum audio energy~\cite{YT-360,ViSAGe} for each clip using spherical harmonics decomposition and adjust the camera's initial orientation so that the trajectory is roughly centered around the primary sound source, as shown in~\cref{fig:teaser}. After obtaining the binaural audio, we discard samples with a normalized left–right channel difference below 0.01, ensuring that the retained samples exhibit distinct spatial auditory cues. 

\subsection{Dataset Statistics}
\label{sec:Statistics}
\DatasetName{} stands out as the largest FoV video dataset with binaural audio format. 
As shown in \cref{tab:dataset}, previous real-world binaural datasets are much smaller in scale. While SimBinaural~\cite{SimBinaural} offers over 100 hours of content, its synthetic video and audio limit its reliability for real-world applications. Moreover, these datasets typically focus on narrow sound environments: OAP~\cite{OAP} on street scenes, and Fair-Play~\cite{Fair-Play}, SimBinaural~\cite{SimBinaural}, and MUSIC-21~\cite{MUSIC1, MUSIC2} on music. In contrast, \DatasetName{} is large-scale and open-domain, offering diverse real-world sound environments. \textit{Spatial Cue Enhancement} (\cref{sec:Enhancement}) further improves spatial perception and dynamics, making it ideal for video-to-binaural spatial audio generation.

%% file: sec/4_method.tex
\input{Figures/network}

\section{\MethodName{} Method}
\label{sec:method}

The overall architecture of \MethodName{} is depicted in Figure~\ref{fig:network}. 
Our approach builds upon parts of the MMAudio~\cite{MMAudio} architecture, fine-tuning the pretrained MMAudio to inherit its robust audio generation capabilities, ensuring strong generalization across open-domain acoustic environments.
To generate binaural audio that ensures both channel coherence and spatial distinctiveness, we adopt \textbf{Dual-Branch Audio Generation} (\cref{sec:dual-branch}), where two dedicated branches independently predict left and right flows.
Additionally, we introduce \textbf{Conditional Spacetime Module} (\cref{sec:spacetime}), which extracts spatio-temporal features from the video and integrates them into the generation process, further enhancing the audio's spatial realism. 

\subsection{Preliminaries}
\label{sec:Preliminaries}
\noindent\textbf{Conditional Flow Matching.} 
We employ the CFM objective~\cite{Flow-Matching, FM2} for audio generation, guiding the model to progressively transform noise $x_0$ sampled from the standard normal distribution into audio latents $x_1$, conditioned on video and text. 
CFM defines a probability density path $p_t(x)$ for $t \in [0,1]$. A common and theoretically grounded choice is to construct this path using Optimal Transport (OT) displacement interpolation~\cite{OT}, yielding:
\begin{equation}
    x_t = t x_1 + (1-t) x_0,
\end{equation}
and the corresponding velocity field at $x_t$ is:
\begin{equation}
    u(x_t \mid x_0, x_1) = x_1 - x_0.
\end{equation}
The goal of CFM is to learn a time-dependent conditional velocity field
$v_\theta(t, \mathbf{C}, x): [0,1] \times \mathbb{R}^C \times \mathbb{R}^d \rightarrow \mathbb{R}^d$, 
parameterized by a neural network with parameters $\theta$, where $t$ is the timestep, $\mathbf{C}$ denotes the conditioning features, and $x$ denotes a point in the vector field.
During training, we optimize $\theta$ using the conditional flow matching objective:
\begin{equation}
    \mathbb{E}_{t, q(x_0), q(x_1,\mathbf{C})} \| v_\theta(t, \mathbf{C}, x_t) - u(x_t \mid x_0, x_1) \|^2,
\end{equation}
where $t$ is uniformly sampled, $q(x_0)$ is the standard normal distribution, and $q(x_1,\mathbf{C})$ is sampled from the training set. 

\noindent\textbf{Binaural Spatial Audio Generation.} 
Given a perspective video $\mathcal{I} = \{ I_n \}^{N}_{n=1}$, where $N$ denotes the number of frames and each $I_n \in \mathbb{R}^{3 \times H \times W}$ is an RGB image, optionally accompanied by text $\mathcal{T}$, our objective is to generate high-quality binaural  spatial audio $\mathcal{A} = \{A^l, A^r\}$, with $l$ and $r$ indicating the left and right channels respectively. The generated audio should be spatially immersive and maintain spatio-temporal consistency with the video. ~\cref{fig:teaser} illustrates our end-to-end pipeline. 

Building on conditional flow matching, we extend the mono flow $v_{\theta}$ into dual learning targets $v_{\theta}^l$ and $v_{\theta}^r$, corresponding to the left and right audio channels. During inference, \MethodName{} estimates flows $v_{\theta}^l, v_{\theta}^r$ for the current latent states $x_t^l, x_t^r$, conditioned on the video and text inputs $\mathcal{C}=\{\mathcal{I}[, \mathcal{T}]\}$, along with timestep $t$. The flows are learned during training by optimizing the following objective:
\begin{equation}
\sum_{a \in \{l, r\}} \mathbb{E}_{t, q(x_0^a), q(x_1^a,\mathcal{C})} \left\| v_\theta^a(t, \mathcal{C}, x_t^a) - (x_1^a - x_0^a) \right\|^2.
\end{equation}
The network then numerically integrates the noises $x_0^l$ and $x_0^r$ over the interval $t \in [0, 1]$, guided by flows $v_{\theta}^l$ and $v_{\theta}^r$. The resulting latents $x_l$ and $x_r$ are separately decoded by the VAE~\cite{mono-vae} into mel spectrograms, which are in turn vocoded~\cite{vocoder} into a binaural audio waveform.

\subsection{Dual-Branch Audio Generation}
\label{sec:dual-branch}
This module centers on the joint generation of the two audio channels.
Although some previous methods~\cite{AudioX, ThinkSound, LAD} utilize stereo audio VAEs directly, these VAEs discard a significant amount of spatial information, merely generating audio in the form of stereo channels without preserving the spatial cues necessary for an immersive auditory experience. 
Instead, our method learns two correlated latent representations, $x_l$ and $x_r$.
These latents remain temporally synchronized and semantically consistent, while each encodes spatial cues unique to its corresponding channel.

\noindent\textbf{Features.}
Multimodal features are extracted during training.
At each timestep $t$, the audio latents $x_t^l$ and $x_t^r$ are derived from the CFM latent space.
We extract multimodal features to capture information at multiple levels from the video.
The text feature $F_{\text{text}}$ and visual feature $F_{\text{vis}}$ are obtained using CLIP~\cite{clip}, which captures coarse-grained semantic content of the video. 
The synchronization feature $F_{\text{sync}}$ is obtained via Synchformer~\cite{Synchformer}, which captures the timing and dynamics of sound events from the video frames.

However, these features do not explicitly encode the spatial positions of sound sources. To address this, we introduce the spatially tuned Perception Encoder~\cite{pe}, which leverages separate learnable positional embeddings for the left and right channels to capture fine-grained spatial and semantic cues from distinct perceptual perspectives, producing the spatial feature 
$F_{\text{pe}}^{\{l,r\}}$ that enables more accurate sound source localization. 

All features are then projected into a shared hidden dimension $h$. After upsampling, 
$\{x_t^l, x_t^r, F_{\text{sync}}, F_{\text{pe}}^{\{l,r\}}\} \in \mathbb{R}^{m \times h}$ 
and 
$\{F_{\text{text}}, F_{\text{vis}}\} \in \mathbb{R}^{h}$,
where $m$ denotes the latent sequence length.
The features are then sequentially passed through $N_1$ \textit{Multimodal Joint Transformer Blocks} and $N_2$ \textit{Single-Modal Branch Transformer Blocks} for cross-modal alignment and channel-specific refinement.

\noindent\textbf{Multimodal Joint Transformer Block.}  
To achieve cross-modal alignment, we build upon the multimodal transformer block design from MMAudio~\cite{MMAudio} and introduce a sequential update mechanism.  
Within each block, the audio latents $x_t^l$ and $x_t^r$ are updated sequentially by the shared block to ensure the two channels remain synchronized and semantically consistent, conditioned on $F_{\text{text}}$, $F_{\text{vis}}$, and $F_{\text{sync}}$, as follows:
\begin{align}
x_t^{l\prime}, F_{\text{vis}}^{\prime}, F_{\text{text}}^{\prime} &= \mathcal{B}_{\text{joint}}\big(x_t^l, F_{\text{vis}}, F_{\text{text}}, F_{\text{sync}}\big), \\
x_t^{r\prime}, F_{\text{vis}}^{\prime\prime}, F_{\text{text}}^{\prime\prime} &= \mathcal{B}_{\text{joint}}\big(x_t^r, F_{\text{vis}}^{\prime}, F_{\text{text}}^{\prime}, F_{\text{sync}}\big).
\end{align}

\noindent\textbf{Single-modal Branch Transformer Block.}  
To further capture channel-specific spatio-temporal details, we incorporate audio-only transformer blocks following FLUX~\cite{FLUX}.  
Each channel is processed independently by its own branch, refining the latents based on channel-specific conditioning:
\begin{align}
x_t^{l\prime} &= \mathcal{B}^l_{\text{single}}\big(x_t^l, F_{\text{sp}}^l\big), \\
x_t^{r\prime} &= \mathcal{B}^r_{\text{single}}\big(x_t^r, F_{\text{sp}}^r\big),
\end{align}
where $\mathcal{B}^l_{\text{single}}$ and $\mathcal{B}^r_{\text{single}}$ denote the single-modal transformer blocks for the left and right branches respectively, while $F_{\text{sp}}$ represents the global spacetime feature extracted from our \textit{Conditional Spacetime Module} (\cref{sec:spacetime}).

\input{Tables/Qualitative}

\subsection{Conditional Spacetime Module}
\label{sec:spacetime}

This module is designed to generate global spacetime features $F_{\text{sp}}$ by incorporating the synchronization feature $F_{\text{sync}}$ and the spatial feature $F_{\text{pe}}$ with global contextual information. This integration improves the spatio-temporal consistency between the generated audio and the video. 

\noindent\textbf{Pe Spatial Feature.} As shown in the left part of~\cref{fig:network}, the spatial feature $F_{\text{pe}}\in \mathbb{R}^{m_{\text{pe}}\times 16\times 16\times h_{\text{pe}}}$ is extracted by the spatially tuned Perception Encoder~\cite{pe}, where $m_{\text{pe}}$ represents the number of video frames (8 fps), $h_{\text{pe}}$ is the feature dimension, and 16 is the patch size. We use PE Spatial because it retains semantic information while producing high-quality spatial features.
$F_{\text{pe}}$ is then downsampled through 2x2 average pooling to reduce spatial dimensions. 
Additionally, two learnable spatial position embeddings, $\mathbf{E}^l$ and $\mathbf{E}^r$, are introduced for the left and right channels, respectively, to capture the spatial characteristics specific to each channel. After projections and upsampling, we obtain the frame-aligned spatial features $F^{a}_{\text{pe}}\in \mathbb{R}^{m \times h}, a\in \{l,r\}$:
\begin{align}
F_{\text{pe}}^{a} = \text{Upsample}(\text{MLP}(\text{Flatten}(\text{Conv}(F_{\text{pe}} +\mathbf{E}^{a})))).
\end{align}

\noindent\textbf{Global Spacetime Feature.} The global spacetime features $F_{\text{sp}}$ serve to inject spatio-temporal information into single-modal branch transformer blocks via adaptive layer normalization (adaLN)~\cite{adaLN}. These features are constructed by integrating the frame-aligned spatial features $F_{\text{pe}}^{\{l,r\}}$ with the synchronization feature $F_{\text{sync}}$ and global conditioning $c_g$:
\begin{align}
F_{\text{sp}}^a = \text{Linear}^a \left( \left[c_g+F_{\text{sync}} ; F_{\text{pe}}^a \right] \right), a \in \{l,r\},
\end{align}
where $\text{Linear}^{\{l,r\}}$ denotes channel-specific linear projections that map the concatenated features from dimension $2h$ to $h$. This design enables the joint representation of spatial layouts and temporal dynamics while preserving distinctive channel-wise characteristics for binaural perception.

%% file: Figures/network.tex
\begin{figure*}
    \centering
    \vspace{-10pt}
    \includegraphics[width=\linewidth]{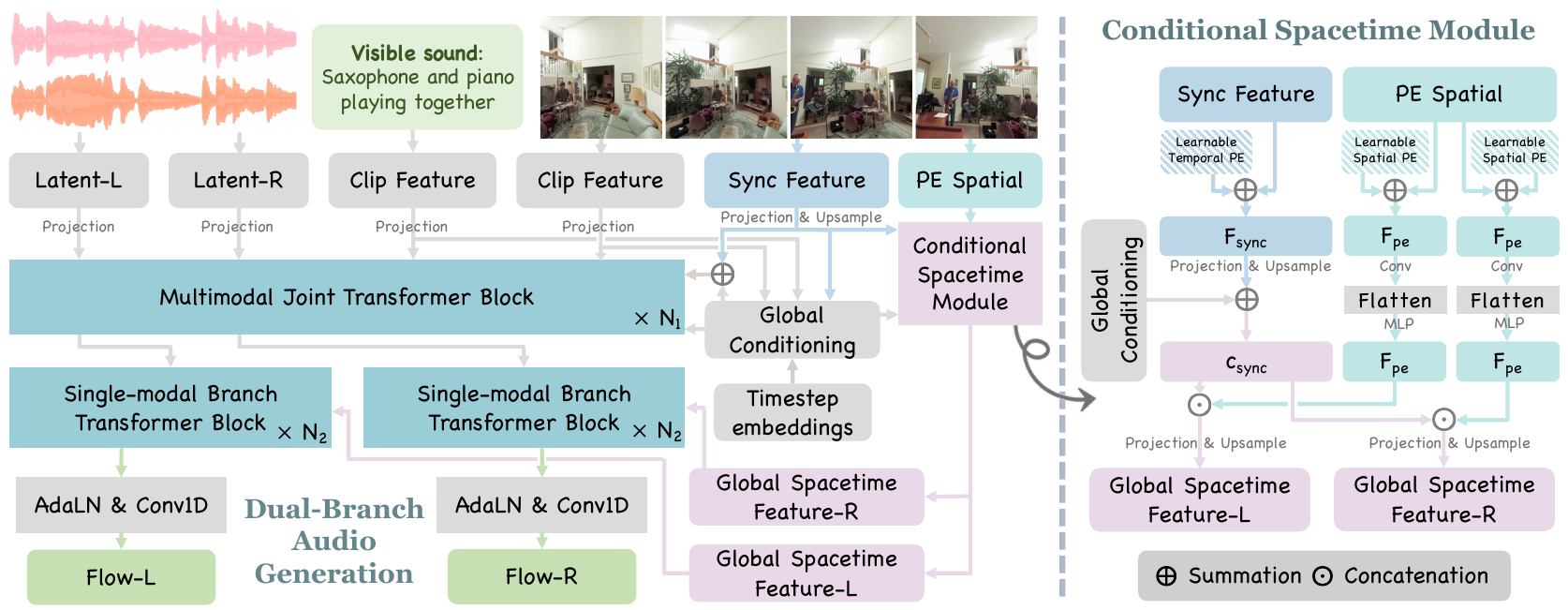}
    \setlength{\abovecaptionskip}{0mm}
    \vspace{-7pt}
    \captionof{figure}{\small \textbf{Our \MethodName{} Network Architecture.} \textbf{Left:} We adopt \textit{Dual-Branch Audio Generation} (\cref{sec:dual-branch}), where two dedicated branches independently predict the left and right audio flows.
    \textbf{Right:} \textit{Conditional Spacetime Module} (\cref{sec:spacetime}) extracts spatiotemporal cues from the video and injects them into the generation process, improving spatio-temporal alignment between audio and video.}
	\label{fig:network}
\end{figure*}

%% file: Tables/Qualitative.tex
\begin{table*}[t]
  \small
  \vspace{-10pt}
    \caption{\textbf{Objective Evaluation Results.} We compare \MethodName{} with baseline methods on the \DatasetName{}, MUSIC, and FAIR-Play test sets. 
    Notably, ThinkSound~\cite{ThinkSound} and AudioX~\cite{AudioX} produce 2-channel stereo audio rather than authentic binaural spatial audio. 
    Comprehensive details regarding the baseline methods and evaluation metrics are provided in \cref{sec:setup}. The best results are highlighted in \textbf{bold}, and the second-best results are \underline{underlined}. Our model consistently outperforms all baselines across all metrics and test sets.}
    \vspace{-6pt}
  \label{tab:AllDatasets}
  \centering
  \begin{tabular}{@{}llcccccccc@{}}
    \toprule
    \multirow{2}{*}{\textbf{Dataset}} &
    \multirow{2}{*}{\textbf{Method}} &
    \multicolumn{6}{c}{\textbf{Audio Distribution Matching}} &
    \multicolumn{2}{c}{\textbf{Video-Audio Alignment}} \\
    \cmidrule(lr){3-8} \cmidrule(lr){9-10}
    & &
    \textbf{$\text{FD}_{\text{VGG}}^{\text{mix}}$$\downarrow$} &
    \textbf{$\text{FD}_{\text{VGG}}^{\text{avg}}$$\downarrow$} &
    \textbf{$\text{FD}_{\text{PANN}}^{\text{mix}}$$\downarrow$} &
    \textbf{$\text{FD}_{\text{PANN}}^{\text{avg}}$$\downarrow$} &
    \textbf{$\text{KL}_{\text{PANN}}^{\text{mix}}$$\downarrow$} &
    \textbf{$\text{KL}_{\text{PANN}}^{\text{avg}}$$\downarrow$} &
    \textbf{DeSync$\downarrow$} &
    \textbf{IB-Score$\uparrow$} \\
    \midrule
    \multicolumn{10}{@{}l}{\textbf{In-distribution}} \\
    \midrule
    \multirow{5}{*}{BiAudio} 
      & ThinkSound~\cite{ThinkSound} & 5.125 & 5.949 & 23.801 & 23.928 & 2.462 & 2.480 & \underline{0.903} & 0.191 \\
      & AudioX~\cite{AudioX} & \underline{4.224} & \underline{3.811} & \underline{22.240} & \underline{20.941} & \underline{2.167} & \underline{2.165} & 1.157 & \underline{0.235} \\
      & ViSAGe~\cite{ViSAGe} & 14.212	& 14.432 & 62.587& 	60.289 &	3.546 &	3.483 &	1.159 &	0.123 \\
      & See2Sound~\cite{See-2-sound} & 9.573 & 9.904 & 42.853 & 41.797 & 3.410 & 3.427 & 1.244 & 0.088 \\
      \rowcolor{lightblue}\cellcolor{white}&   \textbf{\MethodName{} (Ours)} & \textbf{2.516} & \textbf{2.479} & \textbf{13.917} & \textbf{12.684} & \textbf{1.355} & \textbf{1.360} & \textbf{0.788} & \textbf{0.299} \\
    \midrule
    \multirow{5}{*}{MUSIC} 
      & ThinkSound~\cite{ThinkSound} & \underline{16.451} & \underline{16.922} & \underline{14.948} & \underline{14.349} & \underline{0.481} & \underline{0.477} & \underline{0.331} & 0.347 \\
      & AudioX~\cite{AudioX} & 24.299 & 24.173 & 21.602 & 20.992 & 0.689 & 0.695 & 1.322 & \underline{0.350} \\
      & ViSAGe~\cite{ViSAGe} & 48.260 & 50.691 & 104.082 & 100.693 & 3.374 & 3.346 & 1.266 & 0.063 \\
      & See2Sound~\cite{See-2-sound} & 45.038 & 45.912 & 87.253 & 81.747 & 3.247 & 3.111 & 1.269 & 0.067 \\
      \rowcolor{lightblue}\cellcolor{white}& \textbf{\MethodName{} (Ours)} & \textbf{4.649} & \textbf{3.225} & \textbf{8.853} & \textbf{6.288} & \textbf{0.332} & \textbf{0.267} & \textbf{0.178} & \textbf{0.429} \\
    \midrule
    \multicolumn{10}{@{}l}{\textbf{Out-of-distribution}} \\
    \midrule
    \multirow{5}{*}{FAIR-Play} 
      & ThinkSound~\cite{ThinkSound} & \underline{6.648} & \underline{6.232} & \underline{37.207} & \underline{35.811} & \underline{1.916} & \underline{1.951} & \underline{0.764} & 0.118 \\
      & AudioX~\cite{AudioX} & 11.421 & 11.457 & 58.358 & 57.436 & 2.977 & 3.088 & 1.111 & \underline{0.133} \\
      & ViSAGe~\cite{ViSAGe} & 15.010  &	14.660 &	86.568  &	83.241  &	2.514 &	2.468  &	1.103  &	0.067 \\
      & See2Sound~\cite{See-2-sound} & 25.611 & 24.079 & 87.577 & 82.750 & 3.672 & 3.474 & 1.140 & -0.004 \\
      \rowcolor{lightblue}\cellcolor{white}& \textbf{\MethodName{} (Ours)} & \textbf{4.310} & \textbf{4.179} & \textbf{27.134} & \textbf{23.330} & \textbf{1.528} & \textbf{1.538} & \textbf{0.724} & \textbf{0.194} \\
    \bottomrule
  \end{tabular}
\end{table*}

%% file: sec/5_experiment.tex
\section{Experiments}
\label{sec:experiment}

\subsection{Experimental Setup}
\label{sec:setup}
\paragraph{Datasets.}
The binaural audio–video datasets we utilize include our \DatasetName{}, MUSIC-21~\cite{MUSIC1,MUSIC2} and FAIR-Play~\cite{Fair-Play}. Each video is segmented into $8\,\text{s}$ clips, with audio sampled at 44.1 kHz. For all datasets, we discard samples with minimal left-right channel differences, as described in~\cref{sec:dataset}. Subsequently, we split \DatasetName{} by video ID, resulting in 94,845 training clips and 2,695 test clips, and split MUSIC-21 into 17,825 training clips and 1,857 test clips. The FAIR-Play dataset is employed as an out-of-distribution test set, comprising 1,871 clips, to evaluate the model’s generalization ability beyond the training domain.

\paragraph{Baselines.}
The following methods serve as the baselines:

\noindent\textbf{\textit{ThinkSound}}~\cite{ThinkSound} and \textbf{\textit{AudioX}}~\cite{AudioX}: multimodal V2A approaches that leverage stereo VAEs to produce stereo audio, yet ignore spatial information from the video.

\noindent\textbf{\textit{ViSAGe}}~\cite{ViSAGe}: It generates first-order ambisonics (FOA) based on visual content and camera direction. We fix the camera direction at $(44.5, 89.5)$. Since ViSAGe only handles $5\,\text{s}$ video clips, we compress the original $8\,\text{s}$ videos to $5\,\text{s}$, generate the corresponding FOA, render binaural audio, and then restore the output to $8\,\text{s}$.

\noindent\textbf{\textit{See2Sound}}~\cite{See-2-sound}: It generates mono-audio for each visual region of interest and synthesizes 5.1 surround audio. We mix it down to the left and right channels.

\input{Tables/userstudy}

\paragraph{Evaluation Metrics.}
We evaluate \MethodName{} using both objective and subjective metrics to comprehensively assess its performance. The objective metrics are as follows:

\noindent\textit{FD} and \textit{KL}:
We employ Fréchet Distance (FD) with VGGish~\cite{AudioSet} ($\text{FD}_{\text{VGG}}$) and PANN~\cite{PANNs} ($\text{FD}_{\text{PANN}}$) embeddings, as well as Kullback–Leibler divergence (KL) with PANN ($\text{KL}_{\text{PANN}}$) as classifiers, to evaluate the distributional similarity between the generated and ground-truth binaural audio.
To holistically reflect the quality of the generated audio, we report \textbf{\textit{$\text{FD}^{\text{mix}}$} }and \textbf{\textit{$\text{KL}^{\text{mix}}$}} computed on the mixed mono audio. 
To assess the fidelity of the individual channels, we report \textbf{\textit{$\text{FD}^{\text{avg}}$}} and \textbf{\textit{$\text{KL}^{\text{avg}}$}}, which are obtained by averaging the scores from each channel.
We further introduce \textbf{\textit{DeSync}} to evaluate audio-visual synchrony, which measures the temporal misalignment between the generated audio and video, predicted by Synchformer~\cite{Synchformer}, following MMAudio~\cite{MMAudio}.
Moreover, we assess semantic alignment using \textbf{\textit{IB-Score}}, the average cosine similarity between visual features extracted via ImageBind~\cite{ImageBind} and audio features.

However, objective metrics fall short in intuitively assessing the spatial impression and audio-visual consistency of generated audio. Therefore, we conducted a subjective evaluation using Mean Opinion Score (MOS) across five perceptual aspects to incorporate a human-aligned perspective:
\noindent\textbf{\textit{Spatial Impression}}: Whether audio conveys a clear sense of spatiality, including left-right and depth cues.
\noindent \textbf{\textit{Spatial Consistency}}: how well the audio aligns with the visual spatial cues, ensuring that audio and visual elements correspond correctly in 3D space.
\noindent \textbf{\textit{Temporal Alignment}}: Synchronization between audio and the input video, matching the timing of visual events.
\noindent \textbf{\textit{Semantic Alignment}}: Correspondence between the generated binaural spatial audio and the input modalities, including video and optional text.
\noindent \textbf{\textit{Audio Realism}}: Naturalness, clarity, and realism of the generated audio, independent of the video content.

\paragraph{Implementation Details.}
All training experiments are conducted on 8 NVIDIA A800-SXM4-80GB GPUs with 2 AMD EPYC 7H12 64-Core CPUs. We fine-tune the MMAudio~\cite{MMAudio} flow prediction network (large, 44.1kHz, v2) on our training set using a batch size of 64, a learning rate of 1e-4, and a weight decay of 1e-6. Training typically converges after 50,000 iterations, within two days on 8 A800 GPUs. Approximately 4TB of high-speed NVMe storage is required. Details are provided in Appendix~\cref{sec:experiments_settings_sup}.

\subsection{Quantitative Results}
\label{sec:quantitative}
\paragraph{Objective Evaluation.}
We report the superior performance of \MethodName{} across multiple objective metrics on both in-distribution and out-of-distribution datasets, as shown in~\cref{tab:AllDatasets}. It is compared with several state-of-the-art video-to-spatial-audio generation methods, with details of all compared methods provided in~\cref{sec:setup}. For the BiAudio and MUSIC test sets, we use both video and caption inputs, while for FAIR-Play, which lacks captions, only video is provided. Models without text-processing capability are evaluated using video inputs only.
The results in~\cref{tab:AllDatasets} show that \MethodName{} outperforms other binaural-audio generation methods in overall quality, spatial channel distribution matching, audio-visual synchrony, and semantic alignment. This highlights the model’s ability to generate high-fidelity binaural audio while maintaining audio-visual coherence under varying acoustic environments. Notably, its strong performance on the out-of-distribution FAIR-Play dataset, which was not seen during training, demonstrates robust generalization across diverse sound environments.

\noindent\textbf{User Study.}
We performed a comprehensive subjective evaluation in~\cref{tab:UserStudy} to establish human-aligned evaluation metrics. We sampled 10 videos, along with their corresponding captions when available, from three test sets, covering diverse domains, acoustic environments, and events, with both stationary and moving sound sources.
For each video, binaural audio was generated using our method and the baseline approaches. We collected responses from 12 experts. Model performance was evaluated using the Mean Opinion Score (MOS) and its 95\% confidence interval, where participants assigned scores from 1 to 5 to each of the five competing models per task, with higher scores indicating better performance. Each participant rated 50 audio–visual samples according to the five criteria outlined in~\cref{sec:setup}. 
As shown in~\cref{tab:UserStudy}, our approach outperforms all baselines across every metric, closely matching earlier quantitative results and validating both its perceptual quality and technical reliability. Overall, our method achieves state-of-the-art performance in spatial impression, audio-visual consistency, and audio realism.

\subsection{Qualitative Results}
We present a comparative analysis of video-conditioned binaural spatial audio generation in~\cref{fig:qualitative}. The example shows a person playing the sitar, with the camera moving from left to right, causing the sound source to shift from right to left in the video.
We visualize the spectrograms of the generated binaural audio from our method \MethodName{}, the ground truth, and other baselines.
Our method produces audio that is most consistent with the ground truth and effectively reflects spatial changes in the sound source.
The boxed regions indicate that when the sound source is located on the left, the left-channel spectrogram exhibits higher energy. This reflects the increased loudness perceived by the left ear and matches the ground-truth spatial audio, demonstrating our model’s ability to adapt to viewpoint changes. In contrast, other methods generate incorrect rhythms and fail to reflect any left-right channel differences.

\input{Tables/ablation}
\subsection{Ablation Study}

We conduct an ablation study to evaluate the contributions of key model components. We assess three objective metrics to quantify improvements in distribution matching ($\text{FD}_{\text{VGG}}^{\text{avg}}$), temporal alignment (DeSync), and semantic alignment (IB-Score) on our \DatasetName{} test set.  
To further capture the gains in spatial perception, we perform a user study following the setup in~\cref{sec:quantitative}, evaluating Spatial Impression (U-SI) and Spatial Consistency (U-SC) using the Mean Opinion Score (MOS). 
We compare several variants of our model to analyze the impact of different model components: (1) the pretrained MMAudio~\cite{MMAudio} model, spatialized by duplicating the mono output into two channels for evaluation; (2) adding the \textit{Dual-Branch Audio Generation} module to train on our binaural data; and (3) adding the \textit{Conditional Spacetime Module} to integrate spatio-temporal information into the model.  
From the results in~\cref{tab:ablation}, we observe that our \textit{Dual-Branch Audio Generation} module effectively learns the distribution of our binaural training dataset, substantially enhancing spatial generation capabilities. Meanwhile, it retains the pretrained model's strong audio generation quality and generalization, maintaining superior performance in temporal alignment and semantic alignment. 
With the introduction of \textit{Conditional Spacetime Module}, the joint spatio-temporal learning slightly decreases temporal alignment, but further improves the spatial perception of the generated binaural audio. The improvement in U-SC, with a gain of 0.575, demonstrates that the module effectively injects precise spatial information into our architecture, enabling the model to accurately localize sound sources and deliver more reliable and distinguishable spatial cues for binaural audio.

\input{Figures/qualitative}

%% file: Tables/userstudy.tex
\begin{table*}[t]
  \small
  \vspace{-7pt}
    \caption{\textbf{User Study.} Subjective evaluation using Mean Opinion Score (MOS) with 95\% confidence intervals to assess spatial impression, alignment with input video across spatial, temporal, and semantic dimensions, and overall realism of the generated audio.}
    \vspace{-8pt}
  \label{tab:UserStudy}
  \centering
  \begin{tabular}{@{}l|ccccc@{}}
    \toprule
    
    \textbf{Method} &
    \textbf{Spatial Impression}$\uparrow$ &
    \textbf{Spatial Consistency}$\uparrow$ &
    \textbf{Temporal Align.}$\uparrow$ &
    \textbf{Semantic Align.}$\uparrow$ &
    \textbf{Audio Realism}$\uparrow$ \\
    \midrule
     ThinkSound~\cite{ThinkSound} & 3.250  $\pm$ 0.247 & 2.875 $\pm$ 0.294 & 3.483 $\pm$ 0.243 & 3.400 $\pm$ 0.265 & 3.067 $\pm$ 0.272 \\
      AudioX~\cite{AudioX}     & 3.050 $\pm$ 0.272 & 2.792 $\pm$ 0.305 & 2.867 $\pm$ 0.301 & 3.367 $\pm$ 0.261 & 3.258 $\pm$ 0.315 \\
      ViSAGe~\cite{ViSAGe}     & 1.658 $\pm$ 0.228 & 1.517 $\pm$ 0.209 & 1.700 $\pm$ 0.308 & 1.725 $\pm$ 0.310 & 1.492 $\pm$ 0.205 \\
      See2Sound~\cite{See-2-sound}  & 2.033 $\pm$ 0.298 & 1.517 $\pm$ 0.205 & 1.742 $\pm$ 0.333 & 1.725 $\pm$ 0.281 & 1.650 $\pm$ 0.333 \\
    \rowcolor{lightblue}\textbf{ViSAudio (Ours)}  & \textbf{4.133 $\pm$ 0.294} & \textbf{4.100 $\pm$ 0.292} & \textbf{4.275 $\pm$ 0.272} & \textbf{4.292 $\pm$ 0.235} & \textbf{4.158 $\pm$ 0.282} \\
    \bottomrule
  \end{tabular}
\end{table*}
\vspace{-5pt}

%% file: Tables/ablation.tex
\begin{table}[t]
  \centering
  \small
  \setlength{\tabcolsep}{4pt}
  \vspace{-11pt}
\caption{\textbf{Ablation Study on Key Model Components.} We conduct an ablation study to evaluate the contributions of key model components: \textit{Dual, Dual-Branch Audio Generation} (\cref{sec:dual-branch}) and \textit{Spt, Conditional Spacetime Module} (\cref{sec:spacetime}). \textit{Pretrained} refers to our pretrained MMAudio~\cite{MMAudio} model, spatialized by duplicating the mono output into both channels for evaluation.
}
\vspace{-8pt}
  \label{tab:ablation}
  \begin{tabular}{@{}l|ccccc@{}}
  \toprule
    \textbf{Model} & \textbf{$\text{FD}_{\text{VGG}}^{\text{avg}}$$\downarrow$} &
    \textbf{DeSync}$\downarrow$ & \textbf{IB-S}$\uparrow$ & \textbf{U-SI}$\uparrow$ & \textbf{U-SC}$\uparrow$ \\
    \midrule
    Pretrained  & 4.482 & 0.793 & 0.285 & 2.775 & 2.817 \\
    w/ Dual only            & 2.803 & \textbf{0.766} & 0.289 & 4.017 & 3.658 \\
    \rowcolor{lightblue} Dual+Spt          & \textbf{2.479} & 0.788 & \textbf{0.299} & \textbf{4.333} & \textbf{4.233} \\
    \bottomrule
  \end{tabular}
\end{table}

%% file: Figures/qualitative.tex
\begin{figure}
    \centering
    \vspace{-13pt}
    \includegraphics[width=\linewidth]{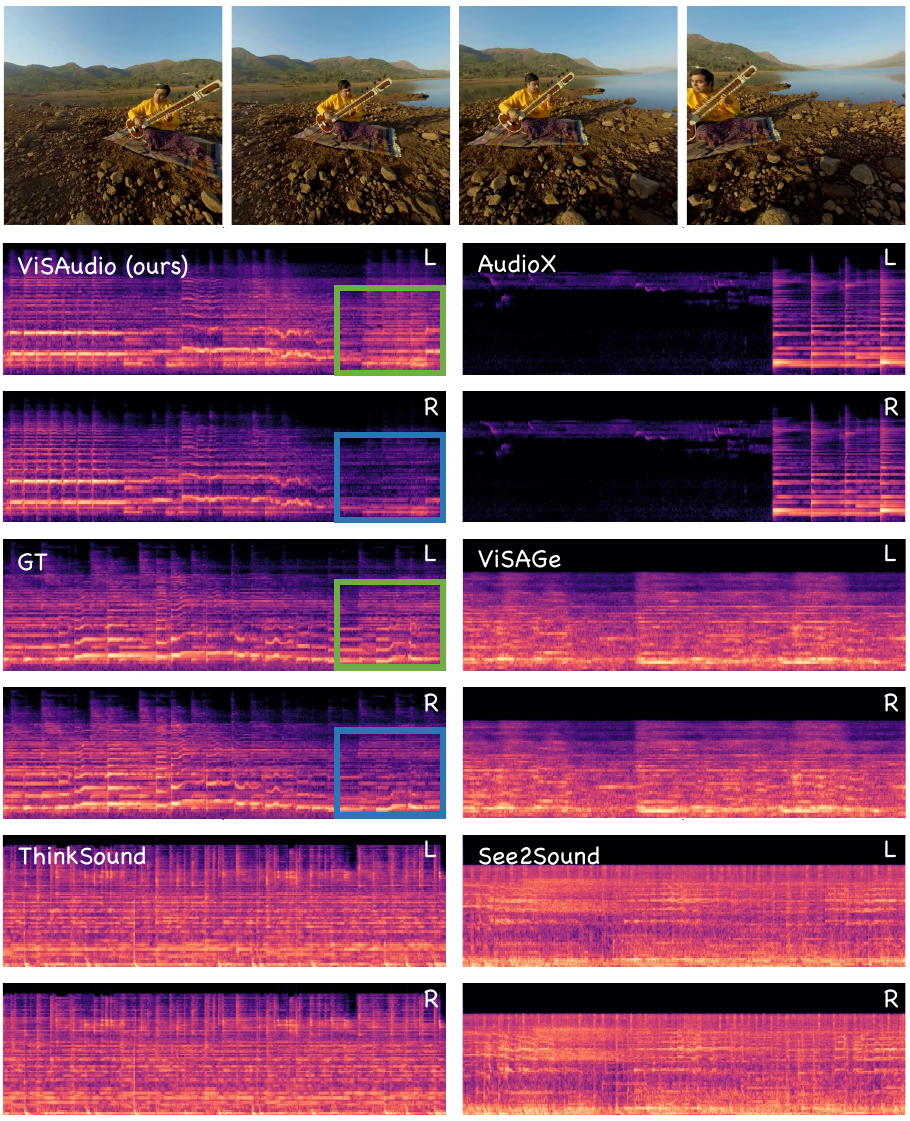}
    \setlength{\abovecaptionskip}{0mm}
    \vspace{-12pt}
    \captionof{figure}{\small 
    \textbf{Qualitative Comparison.}
    The example shows a person playing the sitar while the camera moves from left to right, causing the perceived sound source to shift from right to left. \MethodName{} generates binaural audio that best matches the ground truth and accurately captures the spatial movement of the sound source.
    }
	\label{fig:qualitative}
    \vspace{-6pt}
\end{figure}

%% file: sec/6_conclusion.tex
\section{Conclusion}
\label{sec:conclusion}
We propose \textbf{\MethodName{}}, an end-to-end framework that integrates dual-branch audio generation with a conditional spacetime module to produce spatially immersive binaural audio. To support this task, we curate \textbf{\DatasetName{}}, a large-scale, open-domain dataset of video-binaural pairs featuring diverse camera motions. Our approach achieves state-of-the-art performance, generating high-quality binaural audio with compelling spatial immersion that adapts effectively to viewpoint changes. This work paves the way for more immersive, \textbf{end-to-end} visual-to-spatial-audio generation.

\paragraph{Limitations and future work.} 
Our model currently handles short video clips, which limits its ability to capture long-range temporal dependencies and complex acoustic events. In addition, while \DatasetName{} is an open-domain dataset of real-world recordings, background noise in the audio may introduce artifacts in the generated outputs. Our future work will focus on extending the model to handle longer sequences. 
Looking forward, we plan to extend our framework to support multi-channel audio generation, such as directly producing FOA audio.

%% file: sec/supplementary.tex
The appendix presents supplementary material related to the \textbf{\DatasetName{}} dataset and the \textbf{\MethodName{}} method.
~\cref{sec:dataset_sup} begins with a copyright disclaimer and an overview of the dataset construction pipeline, followed by comprehensive statistics to deepen understanding of its construction and content.
~\cref{sec:method_sup} offers a detailed explanation of the \MethodName{} framework, including the calculation of frame-aligned spatial features.
In~\cref{sec:experiments_sup}, we outline the experimental details, covering model configurations, user study specifics, and additional ablation studies.
Finally, we present cases from the main text and provide further examples through the accompanying video demo. ~\cref{sec:demo} elaborates on the video, showcasing the model's performance across various acoustic scenarios and highlighting its potential for real-world applications.

\section{\DatasetName{} Dataset}
\label{sec:dataset_sup}
\subsection{Disclaimer on Copyright and Data Usage}

The construction of the \DatasetName{} dataset is based on the extension of the existing dataset~\cite{OmniAudio}, and we strictly adhere to the terms of data usage.
The video data utilized in this study were sourced from the YouTube platform. All content is copyrighted by their respective creators and owners. The videos included in this research adhere to YouTube’s terms of service and, where applicable, to Creative Commons licenses. Specifically, videos under the Creative Commons license have been appropriately attributed to the original authors in accordance with the license terms (CC BY 4.0).

For videos not governed by a Creative Commons license, we acknowledge that they are protected by copyright and are used exclusively for academic research purposes. No commercial use of these videos is intended. The use of these videos falls under the fair use doctrine for educational and research purposes, as permitted by copyright law.

\subsection{Dataset Construction Details}
\label{sec:Construction_sup}
We present a semi-automated pipeline that converts raw 360$^\circ$ panoramic video and first-order ambisonics (FOA) audio into field-of-view (FOV) video with dynamic perspectives and corresponding binaural audio. 
The pipeline consists of six key modules:

\noindent\textbf{Sound Source Localization.} 
To enhance perceptually distinct spatial cues, we ensure that primary visual elements corresponding to dominant audio sources remain within the field of view for sustained periods. 

Specifically, we first conduct directional analysis~\cite{YT-360,ViSAGe} of FOA audio through spherical harmonic decomposition to determine the primary sound source direction. In first-order ambisonics (ACN/SN3D), the signal $\mathbf{A}(t) = [W(t), Y(t), Z(t), X(t)]$ consists of four components: $W$, $X$, $Y$, and $Z$. $W$ represents the omnidirectional component, capturing the overall sound pressure without any directional bias, essentially the total sound intensity from all directions. The directional components, $X$, $Y$, and $Z$, correspond to different spatial dimensions of the sound field: $X$ captures the front-back direction, $Y$ represents the left-right direction, and $Z$ corresponds to the up-down direction. These components are derived from spherical harmonics, with $W$ being the zero-order component and $X$, $Y$, and $Z$ being the first-order components, which together describe the full 3D sound field. We compute the energy distribution of the sound field through spherical harmonics decomposition:
\begin{equation}
E(\phi, \theta) = \frac{1}{T} \int_{0}^{T} \left| \sum_{l=0}^{1} \sum_{m=-l}^{l} a_{lm}(t) Y_{l}^{m}(\phi, \theta) \right|^2 dt,
\end{equation}
where $Y_{l}^{m}(\phi, \theta)$ denotes the real spherical harmonic functions of order $l$ and degree $m$, $\phi \in [-\pi,\pi]$ represents the azimuth angle, and $\theta \in [-\pi/2,\pi/2]$ represents the elevation angle. The spherical harmonic functions for first-order ambisonics are defined as:
\begin{align}
Y_0^0(\phi, \theta) &= 1, \\
Y_1^{-1}(\phi, \theta) &= \sin\phi\cos\theta, \\
Y_1^0(\phi, \theta) &= \sin\theta, \\
Y_1^1(\phi, \theta) &= \cos\phi\cos\theta.
\end{align}
The coefficients $a_{mn}(t)$ are derived from the ambisonics components through the relationship:
\begin{align}
a_{00}(t) &= W(t), \\
a_{1,-1}(t) &= Y(t), \\
a_{1,0}(t) &= Z(t), \\
a_{1,1}(t) &= X(t).
\end{align}
The energy distribution map $E(\phi, \theta)$ is computed over a discrete spherical grid with angular resolution of $2^\circ$. The primary sound source direction $(\phi_{\text{max}}, \theta_{\text{max}})$ is determined by locating the position of maximum energy:
\begin{equation}
(\phi_{\text{max}}, \theta_{\text{max}}) = \underset{\phi, \theta}{\mathrm{argmax}} \ E(\phi, \theta).
\end{equation}

\input{Figures_supp/llm_caption}

\noindent\textbf{Dynamic Viewpoint Trajectory Generation.} To overcome the limitations of fixed viewing perspectives, we introduce dynamic camera rotation trajectories. After localizing the primary sound source direction $(\phi_{\text{max}}, \theta_{\text{max}})$, the camera's initial orientation is adjusted to positions near this direction, followed by a gradual drift to simulate natural, fluid movements. This approach ensures that the viewing trajectory evolves around the primary sound source region, keeping the main visual elements corresponding to the audio within the field of view for extended periods.

The camera rotation trajectory is parameterized by three Euler angles: yaw $\phi(t)$, pitch $\theta(t)$, and roll $\psi(t)$. The trajectory generation follows a piecewise linear model:
\begin{align}
\phi(t) &= \phi_0 + \alpha_{\phi} \cdot t,\\
\theta(t) &= \theta_0 + \alpha_{\theta} \cdot t, \\
\psi(t) &= \psi_0 + \alpha_{\psi} \cdot t,
\end{align}
where $t \in [0, T]$ denotes the temporal coordinate, and $(\phi_0, \theta_0, \psi_0)$ represent the initial Euler angles. The initial yaw angle $\phi_0$ is determined by:
\begin{align}
\phi_0 &= \phi_{\text{max}} + \Delta \phi,\\
\Delta \phi &\sim 
\begin{cases}
\mathcal{U}\left(-\frac{\pi}{4}, 0\right) & \text{if } \alpha_{\phi} \cdot T > \frac{\pi}{6}, \\
\mathcal{U}\left(0, \frac{\pi}{4}\right) & \text{if } \alpha_{\phi} \cdot T < -\frac{\pi}{6}, \\
\mathcal{U}\left(-\frac{\pi}{6}, \frac{\pi}{6}\right) & \text{otherwise}.
\end{cases}
\end{align}
The initial pitch angle $\theta_0$ is computed as:
\begin{align}
\theta_0 &= \text{clip}\left(\Delta\theta, -\frac{\pi}{6}, \frac{\pi}{6}\right), \\
\Delta\theta &\sim
\begin{cases}
\mathcal{U}(0, \theta_{\text{max}}) & \text{if } \theta_{\text{max}} > 0, \\
\mathcal{U}(\theta_{\text{max}}, 0) & \text{otherwise}.
\end{cases}
\end{align}
The initial roll angle is fixed to $\psi_0 = 0$.

The drift coefficients $(\alpha_{\phi}, \alpha_{\theta}, \alpha_{\psi})$ are randomly sampled from a uniform distribution with a certain probability of being zero, to simulate diverse motion patterns:
\begin{align}
\alpha_{\phi} &\sim \mathcal{U}\left(-\alpha_{\phi}^{\text{max}}, \alpha_{\phi}^{\text{max}}\right) \cdot \mathbb{I}(\xi_{\phi} > \frac{1}{3}), \\
\alpha_{\theta} &\sim \mathcal{U}\left(-\alpha_{\theta}^{\text{max}}, \alpha_{\theta}^{\text{max}}\right) \cdot \mathbb{I}(\xi_{\theta} > \frac{1}{3}), \\
\alpha_{\psi} &\sim \mathcal{U}\left(-\alpha_{\psi}^{\text{max}}, \alpha_{\psi}^{\text{max}}\right) \cdot \mathbb{I}(\xi_{\psi} > \frac{1}{3}),
\end{align}
where $\mathcal{U}(a, b)$ denotes a uniform distribution in the range $[a, b]$, and $\mathbb{I}(\xi > \frac{1}{3})$ is an indicator function that randomly zeroes out the drift coefficient with a probability of $\frac{1}{3}$. The values $\xi_{\phi}$, $\xi_{\theta}$, and $\xi_{\psi}$ are independent random variables uniformly distributed between 0 and 1. The maximum drift rates are set as follows: $\alpha_{\phi}^{\text{max}} = \frac{\pi}{18}$ for yaw, $\alpha_{\theta}^{\text{max}} = \frac{\pi}{90}$ for pitch, and $\alpha_{\psi}^{\text{max}} = \frac{\pi}{180}$ for roll.

\input{Figures_supp/vocabulary}

\noindent\textbf{Video Rendering.} 
Based on the generated viewpoint trajectory, we convert the 360$^\circ$ equirectangular projection video into field-of-view (FOV) perspective sequences using the method described in~\cite{pyequilib2021github}. For each frame at time $t$, the corresponding rotation matrix $\mathbf{R}(t) = \mathbf{R}_x(\psi(t)) \mathbf{R}_y(-\theta(t)) \mathbf{R}_z(\phi(t))$ is applied to transform the viewpoint. The composite rotation matrix is constructed from elementary rotations about the cardinal axes:
\begin{align}
\mathbf{R}_z(\phi) &= \begin{bmatrix}
\cos\phi & -\sin\phi & 0 \\
\sin\phi & \cos\phi & 0 \\
0 & 0 & 1
\end{bmatrix}, \\
\mathbf{R}_y(\theta) &= \begin{bmatrix}
\cos\theta & 0 & \sin\theta \\
0 & 1 & 0 \\
-\sin\theta & 0 & \cos\theta
\end{bmatrix}, \\
\mathbf{R}_x(\psi) &= \begin{bmatrix}
1 & 0 & 0 \\
0 & \cos\psi & -\sin\psi \\
0 & \sin\psi & \cos\psi
\end{bmatrix}.
\end{align}
The rendering process employs a fixed field of view $\text{FOV} = 90^\circ$ and outputs a video at a resolution of $512 \times 512$ pixels, ensuring the creation of natural perspective sequences that dynamically follow the camera trajectory.

\noindent\textbf{Audio Spatialization.} 
The FOA audio stream undergoes frame-synchronized spatial transformation aligned with the viewing trajectory, followed by binaural rendering through a multi-stage processing pipeline. The audio is divided into $M$ temporal segments corresponding to the trajectory points, with each segment containing $N/M$ samples, where $N$ denotes the total number of audio samples in the stream.
For the $i$-th segment spanning frames $t_i$ to $t_{i+1}$, we construct the rotation matrix $\mathbf{R}_i = \mathbf{R}_x(\psi_i) \mathbf{R}_y(-\theta_i) \mathbf{R}_z(\phi_i)$ using the corresponding Euler angles $(\phi_i, \theta_i, \psi_i)$ from the viewpoint trajectory.
The directional components within each segment are rearranged from the ACN channel order $[W_i, Y_i, Z_i, X_i]$ and spatially transformed according to:
\begin{equation}
    \begin{bmatrix} X^{\text{rot}}_{i} \\ Y^{\text{rot}}_{i} \\ Z^{\text{rot}}_{i} \end{bmatrix} =  \begin{bmatrix} X_i \\ Y_i \\ Z_i \end{bmatrix} \mathbf{R}_i^\top,
\end{equation}
where the omnidirectional component $W_i$ remains unchanged throughout the rotation process. This segment-wise transformation ensures continuous spatial alignment between the audio field and the dynamically evolving camera perspective. The final output maintains the ambisonics signal in the channel order $[W^{\text{rot}}, Y^{\text{rot}}, Z^{\text{rot}}, X^{\text{rot}}]$, consistent with the original ACN layout.

\input{Figures_supp/traj}

The spatially rotated ambisonics signal is rendered to binaural audio through convolution with head-related impulse responses (HRIRs) from the Omnitone library~\cite{Omnitone}. The ambisonics channels are grouped into two pairs and processed through partitioned convolution as follows:
\begin{align}
\mathbf{wy}_{\text{conv}} &= \text{conv}\left(\begin{bmatrix} W \\ Y \end{bmatrix}, \mathbf{hrir}_{wy}\right), \\
\mathbf{zx}_{\text{conv}} &= \text{conv}\left(\begin{bmatrix} Z \\ X \end{bmatrix}, \mathbf{hrir}_{zx}\right),
\end{align}
where $\text{conv}(\cdot)$ denotes the full linear convolution operation, and $\mathbf{hrir}_{wy}$, $\mathbf{hrir}_{zx}$ are the first-order FOA HRIRs decoded from Omnitone~\cite{Omnitone}. The binaural output signals are then synthesized by combining the convolved components:
\begin{align}
L &= \mathbf{wy}_{\text{conv}}[0] + \mathbf{wy}_{\text{conv}}[1] + \mathbf{zx}_{\text{conv}}[0] + \mathbf{zx}_{\text{conv}}[1], \\
R &= \mathbf{wy}_{\text{conv}}[0] - \mathbf{wy}_{\text{conv}}[1] + \mathbf{zx}_{\text{conv}}[0] + \mathbf{zx}_{\text{conv}}[1].
\end{align}

\noindent\textbf{Caption Generation.} To incorporate semantic guidance into our framework, we generate descriptive captions that go beyond basic sound tagging by accounting for the complexities of real-world auditory perception. Human hearing extends beyond the immediate field of view, encompassing both localized on-screen sounds and non-localizable audio elements such as off-screen sources and ambient environmental noise. As illustrated on the left of \cref{fig:llm_caption}, viewers can accurately spatialize visible sound sources like \textit{water splashing}, but cannot precisely localize off-screen sounds such as \textit{birds chirping}. The direct use of such unprocessed audio-visual data may introduce semantic interference and compromise audio-visual alignment.

To address this issue, we design a two-stage captioning pipeline that precisely distinguishes visible and non-visible sound sources, as illustrated in~\cref{fig:llm_caption}. In the first stage, given the perspective video and its corresponding audio as input, \textit{Qwen2.5-Omni}~\cite{Qwen2.5-Omni} produces comprehensive descriptions that capture both visible audio events and background sound elements, including off-screen sources and ambient noise. In the second stage, these detailed descriptions are refined by \textit{Qwen3-Instruct-2507}~\cite{qwen3} into structured captions.

\noindent\textbf{Dataset Filtering.}
To ensure reliable spatial supervision, we apply a filtering procedure to the collected binaural audio data. Given a binaural waveform, we first normalize each channel by removing its DC component and scaling it to the range $[-1, 1]$:
\begin{equation}
    \mathbf{x}_a \leftarrow \frac{\mathbf{x}_a - \mu_a}{\max(|\mathbf{x}_a - \mu_a|) + \epsilon}, \quad a \in \{l, r\},
\end{equation}
where $\mu_a$ denotes the mean of channel $a$, and $\epsilon = 10^{-9}$ avoids numerical instability. We then evaluate the spatial distinctiveness of each audio clip by measuring the mean absolute difference between the two normalized channels:
\begin{equation}
D = \frac{1}{T} \sum_{t=1}^{T} \left| x_l(t) - x_r(t) \right|.
\end{equation}
Only samples satisfying $D > \tau$ with $\tau = 0.01$ are retained. This filtering criterion ensures that the curated dataset contains audio clips with sufficiently strong inter-channel disparities, which are essential for learning robust and meaningful spatial auditory representations.

\subsection{Dataset Statistics Details}
In this section, we provide detailed statistics of our dataset, covering the vocabulary, initial positions of viewpoint, and viewpoint trajectory visibility.

\noindent \textbf{Vocabulary:} The dataset includes 2,360 unique sound categories or descriptive terms for visible sounds and 1,265 for invisible sounds. The distribution of these categories is illustrated in~\cref{fig:vocabulary}, with the left panel showing the top 50 nouns in bar charts and the right panel displaying the top 200 nouns in the vocabulary as word clouds. These statistics indicate that our dataset is open-domain and highly diverse, covering a wide range of acoustic environments.

\noindent \textbf{Initial Camera Viewpoint:} We analyze the spatial distribution of the camera viewpoint at the start of each clip. The heatmap in~\cref{fig:traj_a} shows the distribution of yaw and pitch angles, highlighting which regions of the spherical field of view are most commonly observed.

\noindent \textbf{Camera Rotation Trajectory:} We report statistics on how the camera rotates along the viewpoint trajectories. ~\cref{fig:traj_b} visualizes 500 sampled 3D rotation trajectories on a unit sphere, where each curve represents a temporally evolving viewing direction. The color along each curve transitions from light to dark, indicating progression through time, which allows us to observe how the camera moves across the field of view during the clip.

\section{GenDoP Method}
\label{sec:method_sup}
\subsection{Frame-aligned Spatial Features}

\input{Tables_supp/algo}

In this section, we provide a detailed explanation of how the \textit{frame-aligned spatial features} mentioned in Section 4.3 (\textit{Conditional Spacetime Module}, \textit{PE Spatial Feature}) are obtained. The frame-aligned features 
$F_{\text{pe}}^{a} \in \mathbb{R}^{m \times h}$, for $a \in \{l,r\}$, are computed as:
\begin{equation}
F_{\text{pe}}^{a} = \text{Upsample}\Big(\text{MLP}\big(\text{Flatten}(\text{Conv}(F_{\text{pe}} + \mathbf{E}^{a}))\big)\Big).
\end{equation}
The detailed processing steps are presented in~\cref{alg:pe_input_proj}.

The spatial feature 
$F_{\text{pe}} \in \mathbb{R}^{m_{\text{pe}} \times 16 \times 16 \times h_{\text{pe}}}$ 
is extracted using the spatially tuned Perception Encoder~\cite{pe}, where $m_{\text{pe}}$ denotes the number of video frames (8 fps), $h_{\text{pe}}$ is the feature dimension, and $16$ corresponds to the patch size. 
Two learnable spatial position embeddings, $\mathbf{E}^l$ and $\mathbf{E}^r$, are introduced for the left and right audio channels, respectively, to capture channel-specific spatial characteristics. 
Subsequently, a series of convolutional and linear projections are applied to transform these features into frame-aligned features 
$F^a_{\mathrm{pe}} \in \mathbb{R}^{m \times h}, a \in \{l,r\}$, 
where $m$ corresponds to the target audio sequence length and $h$ is the hidden feature dimension. 
These frame-aligned features can then be merged with the synchronization feature $F_{\text{sync}}$ to construct the global spacetime features $F_{\text{sp}}$, which serve to inject spatio-temporal information into the subsequent audio generation modules.

\section{Experiments}
\label{sec:experiments_sup}
\subsection{Experimental Settings}
\label{sec:experiments_settings_sup}
\noindent \textbf{Model Configuration.} 
All experiments are conducted by fine-tuning the \texttt{large}, \texttt{44.1kHz}, \texttt{v2} variant of MMAudio~\cite{MMAudio}.  
The model uses a latent dimension of 40, with feature dimensions of 1024 for clip features, 768 for synchronization features, 1024 for PE spatial features, and 1024 for text embeddings. Input sequence lengths are 345 tokens for audio latent representations, 64 frames for video clip sequences, 192 for synchronization sequences, 64 for PE spatial feature sequences, and 77 tokens for text captions. Video clip and text sequences are averaged over the temporal dimension to yield a single conditional feature vector, while other features are temporally aligned prior to fusion. A patch size of 16 is applied during PE spatial feature extraction.
The transformer backbone comprises 7 layers of \textit{multimodal joint transformer blocks} and 14 layers (28 blocks) of \textit{single-modal branch transformer blocks}, with a hidden dimension of 896 with 14 attention heads.
Using the notation introduced in the preceding section:
$
m_{\mathrm{pe}} = 64,  h_{\mathrm{pe}} = 1024,  m = 345,  h = 896.
$

\noindent \textbf{Training Details.}
In all experiments, the model is trained on a combined dataset comprising BiAudio and MUSIC~\cite{MUSIC1, MUSIC2}. Specifically, the BiAudio training set is loaded along with the MUSIC training set, which is oversampled three times to balance the dataset. For the MUSIC dataset, textual captions are automatically generated by parsing instrument names from the corresponding audio file names. For instance, the video file \textit{duet\_acoustic\_guitar\_violin\_WeeRb3LMb8E.mp4} yields the caption \textit{``acoustic guitar, violin''}. 

\noindent \textbf{Inference Details.}
For the BiAudio dataset, during inference, we remove the \textit{Visible} and \textit{Invisible} labels used during training, retaining only the descriptions of the sounds to ensure a fair comparison across methods.

\input{Figures_supp/user_study}
\input{Tables_supp/ablation_module}
\input{Tables_supp/ablation_dataset}

\subsection{User Study Details}

The user study uses a web-based questionnaire. Participants wear headphones on both ears to ensure proper spatial audio perception. 
~\cref{fig:userstudy_a} shows the subjective evaluation criteria presented to participants. It includes the five perceptual dimensions, \textit{Spatial Impression}, \textit{Spatial Consistency}, \textit{Temporal Alignment}, \textit{Semantic Alignment}, and \textit{Audio Realism}, along with their specific scoring guidelines. ~\cref{fig:userstudy_b} depicts the rating interface where participants evaluate five spatial audio samples within each video–audio group. For each sample, participants provide ratings from 1 to 5 for all five perceptual dimensions.

\subsection{Additional Ablation Studies}
In this section, we present additional ablation studies to further investigate the contributions of different model components in our \MethodName{} model and the \DatasetName{} dataset.

\cref{tab:ablation_module} presents the results of an ablation study on key model components. The table provides the complete objective metric results, serving as a supplement to~\cref{tab:ablation} in the main text. The findings highlight that the combination of both modules significantly improves performance.

Additionally, we conduct an ablation study to evaluate the impact of our \DatasetName{} dataset. We compare several variants of our model: (1) the model trained without \DatasetName{}, using only the MUSIC dataset~\cite{MUSIC1, MUSIC2}; (2) the model trained on both \DatasetName{} and MUSIC. As shown in \cref{tab:ablation_dataset}, the model trained with \DatasetName{} performs significantly better on open-domain data, underscoring its importance in generating high-quality binaural spatial audio.

\section{Additional Cases}
\label{sec:demo}

A supplementary video is provided, showcasing cases from the main text along with additional binaural spatial audio cases generated in diverse acoustic environments from input video and optional text prompts. The audiovisual presentation allows the audience to better evaluate the model's ability to produce realistic and spatially precise sound. The video is organized into the following sections for detailed demonstration:

\subsection{Dynamic Sound Sources}
We evaluate the model under diverse motion configurations to assess its spatiotemporal modeling capabilities. With a fixed viewpoint, the model accurately localizes static objects and reliably tracks moving sound sources. Under dynamic camera motion, it maintains stable spatial perception of stationary sound sources. These experiments collectively demonstrate the model's robustness in handling complex auditory scenarios involving both object and viewpoint dynamics, generating high-quality binaural audio with spatial immersion that adapts seamlessly to \textit{viewpoint changes} and \textit{sound-source motion}.

\subsection{Multiple Sound Sources}
We showcase the model's capacity to handle multiple simultaneous sounds across three scenarios: identical sources at different positions, demonstrating precise spatial discrimination; two interacting sources, showing effective separation and localization; and complex ensembles with overlapping sources. 
These examples collectively validate the model's ability to generate coherent auditory scenes with accurate spatial perception in \textit{complex sound environments}.

\subsection{Invisible Sound Sources}
Our method supports both video-only and video-text multimodal conditioning. In video-only mode, the model relies exclusively on visual frames to infer audio. When text captions are provided, they deliver explicit semantic guidance, enabling the generation of off-screen sounds. We present examples where sounds originate from off-screen sources, demonstrating the model's capacity to infer and spatialize unseen audio events based on contextual cues.

\subsection{Diverse Acoustic Environments}
The examples illustrate the model’s strong generalization ability across a range of acoustic environments, including outdoor, underwater, and indoor scenes. It consistently produces high-quality binaural audio with convincing spatial immersion, validating its robustness in adapting to \textit{diverse acoustic environments}.

\subsection{Cases in the Main Text}
The video also includes demonstrations of the results contained in the main text, namely the two cases from~\cref{fig:teaser} and the case from~\cref{fig:qualitative}.

%% file: Figures_supp/llm_caption.tex
\begin{figure}
    \centering
    \includegraphics[width=\linewidth]{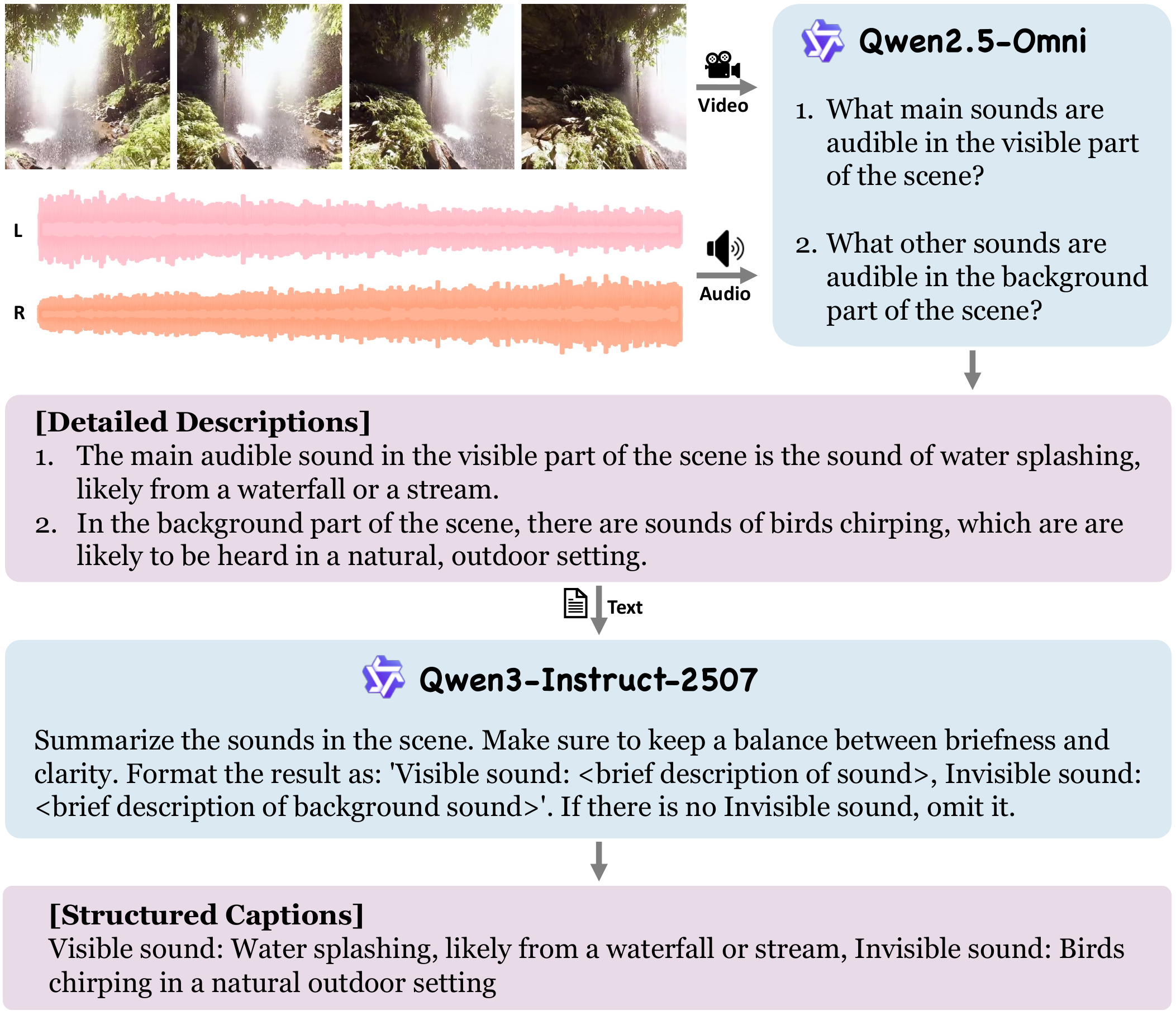}
    \setlength{\abovecaptionskip}{0mm}
    \vspace{-7pt}
    \caption{\textbf{Caption Annotation Pipeline.} We design a two-stage annotation pipeline to label visible and non-visible sound sources. First, \textit{Qwen2.5-Omni}~\cite{Qwen2.5-Omni} generates comprehensive textual descriptions that capture both visible sounds and background audio elements, including off-screen sources and environmental noise. These detailed descriptions are subsequently refined by \textit{Qwen3-Instruct-2507}~\cite{qwen3} into structured captions.}
	\label{fig:llm_caption}
\end{figure}

%% file: Figures_supp/vocabulary.tex
\begin{figure*}
    \centering
    \includegraphics[width=\textwidth]{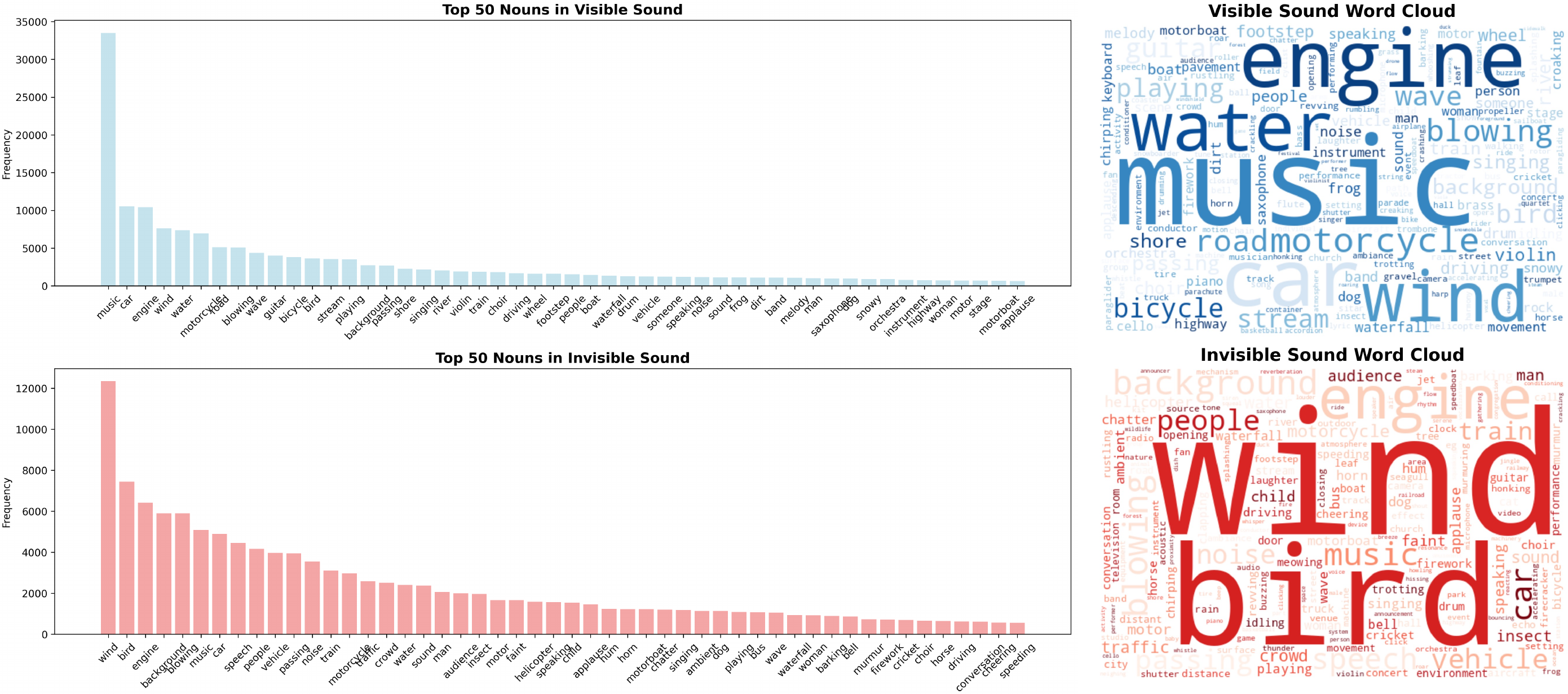}
    \vspace{-20pt}
    \caption{\textbf{Vocabulary in BiAudio captions.} \textbf{Left:} Bar charts displaying the top 50 nouns for visible (blue) and invisible (red) sound sources. \textbf{Right:} Word clouds illustrating the distribution of the top 200 nouns in the vocabulary.}
    \label{fig:vocabulary}
    \vspace{-5pt}
\end{figure*}

%% file: Figures_supp/traj.tex
\begin{figure*}[t]
    \centering
    \begin{subfigure}[b]{0.58\textwidth}
        \centering
        \includegraphics[width=\textwidth]{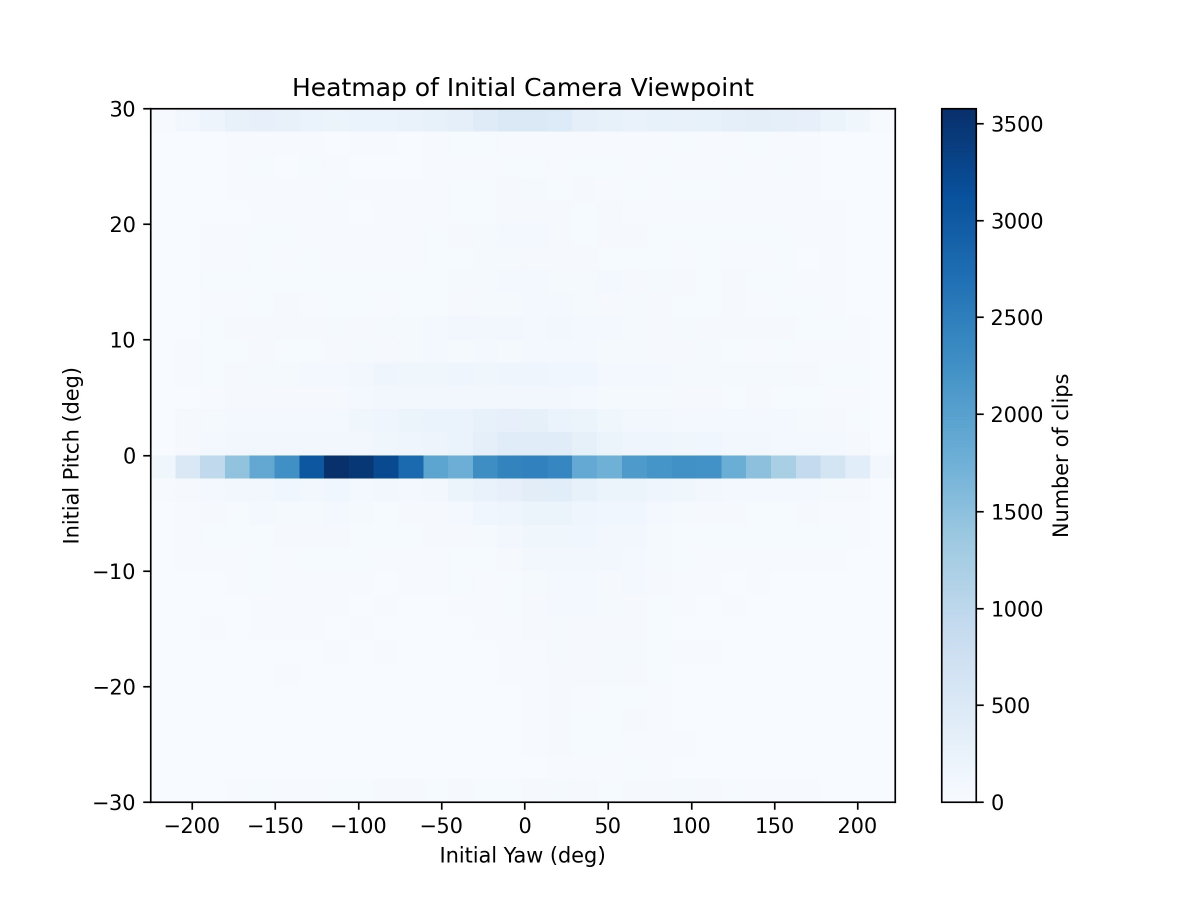}
        \caption{Heatmap of initial camera viewpoints.}
        \label{fig:traj_a}
    \end{subfigure}
    \hfill
    \begin{subfigure}[b]{0.41\textwidth}
        \centering
        \includegraphics[width=\textwidth]{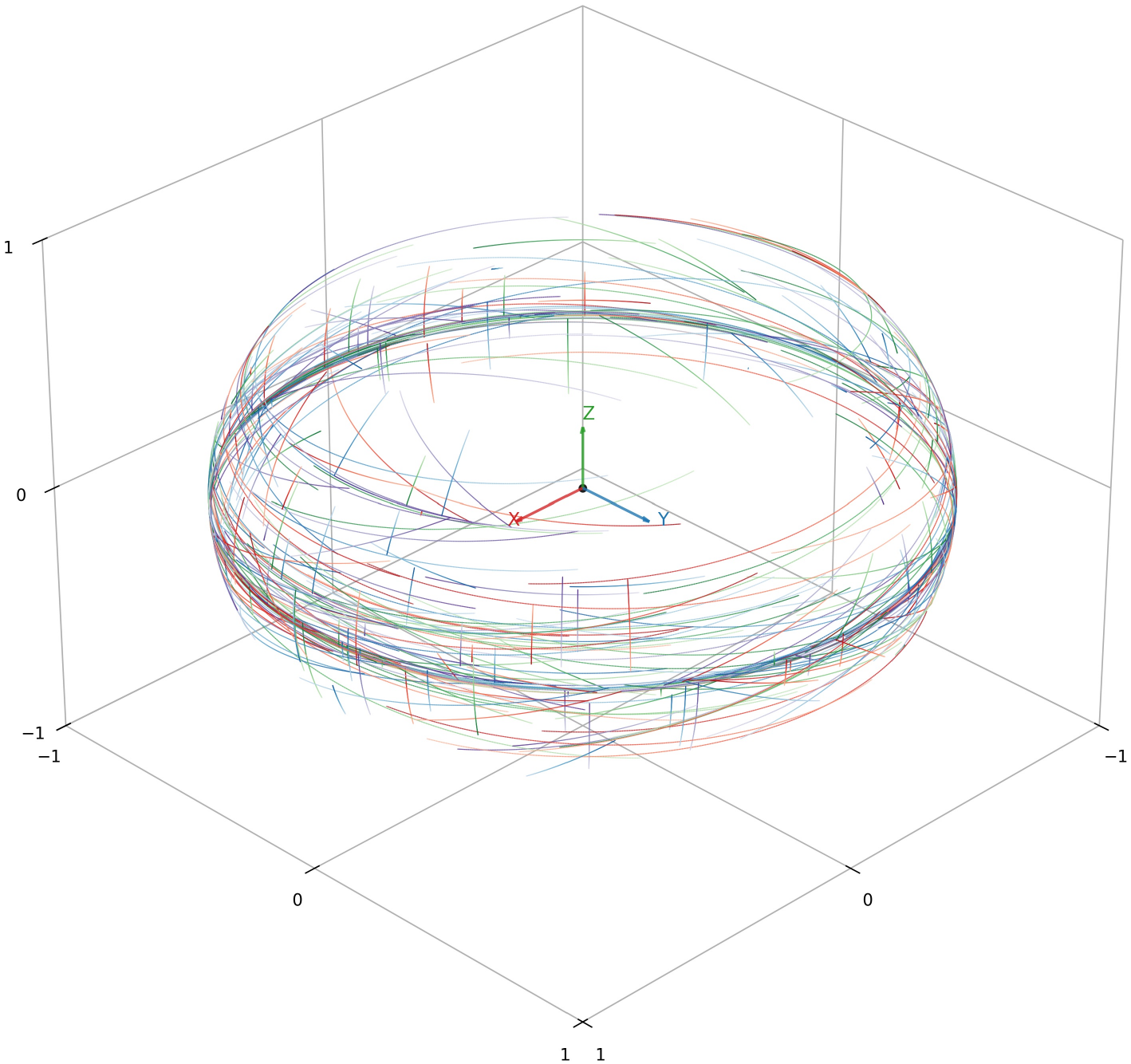}
        \caption{3D visualization of camera rotation trajectories}
        \label{fig:traj_b}
    \end{subfigure}
    \vspace{-5pt}
    \caption{\textbf{Camera Trajectory Statistics in BiAudio.}
    \textbf{(a)} Heatmap of initial camera viewpoints, showing the distribution of starting yaw--pitch angles across all clips.
    \textbf{(b)} 3D visualization of camera rotation trajectories on a unit sphere. Each curve represents a temporally evolving viewing direction, with color changing from light to dark to indicate progression along the trajectory.}
    \label{fig:traj}
    \vspace{-5pt}
\end{figure*}

%% file: Tables_supp/algo.tex
\begin{algorithm}[t]
\caption{Frame-Aligned PE Feature Projection}
\small
\label{alg:pe_input_proj}
\begin{algorithmic}[1]
\Statex \hspace{-1em}\textbf{Input:} 
 PE features $F_{\mathrm{pe}} \in \mathbb{R}^{b \times m_{pe} \times H \times W \times h_{pe}}$, 
\Statex \quad batch size $b$, frames $m_{pe}$, height $H$, width $W$, channels $h_{pe}$
\Statex \hspace{-1em}\textbf{Output:} 
 Frame-aligned features $F^a_{\mathrm{pe}} \in \mathbb{R}^{b \times m \times h}, a \in \{l,r\}$
\Statex \quad target audio length $m$, hidden dimension $h$

\State \textbf{Add learnable stereo position embeddings:} 
\[
F^a_{\mathrm{pe}} \gets F_{\mathrm{pe}} + \mathbf{E}_{\mathrm{pos}}^a
\]

\State \textbf{Channel-wise 1D convolution:} 
\[
F^a_{\mathrm{pe}} \gets \mathrm{SiLU}(\mathrm{Conv1d}(F^a_{\mathrm{pe}}))
\]  

\State \textbf{Spatial downsampling via strided convolution:} 
\[
F^a_{\mathrm{pe}} \gets \mathrm{SiLU}(\mathrm{Conv2d}(F^a_{\mathrm{pe}}, \mathrm{stride}=2)) \quad \text{// $H/2 \times W/2$}
\]  
\[
F^a_{\mathrm{pe}} \gets \mathrm{SiLU}(\mathrm{Conv2d}(F^a_{\mathrm{pe}}, \mathrm{stride}=2)) \quad \text{// $H/4 \times W/4$}
\]

\State \textbf{Flatten spatial dimensions:} 
\[
F^a_{\mathrm{pe}} \gets \text{reshape}(F^a_{\mathrm{pe}}, (b \cdot m_{pe}, (H/4) \cdot (W/4), h))
\]

\State \textbf{Gated convolution (ConvMLP):} 
\[
F^a_{\mathrm{pe}} \gets \mathrm{Conv1d}_2(\mathrm{SiLU}(\mathrm{Conv1d}_1(F^a_{\mathrm{pe}})) \circ \mathrm{Conv1d}_3(F^a_{\mathrm{pe}}))
\]

\State \textbf{Two-layer linear projection:} 
\[
F^a_{\mathrm{pe}} \gets \mathrm{Linear}(\mathrm{SiLU}(\mathrm{Linear}(F^a_{\mathrm{pe}})))
\]  

\State \textbf{Restore batch and temporal dimensions:} 
\[
F^a_{\mathrm{pe}} \gets \text{reshape}(F^a_{\mathrm{pe}}, (b, m_{pe}, h))
\]

\State \textbf{Temporal upsampling to audio length:}
\[
F^a_{\mathrm{pe}} \gets \text{Interpolate}(F^a_{\mathrm{pe}}, \text{size}=m, \text{mode}=linear)
\]  
\State \Return $F^a_{\mathrm{pe}} \in \mathbb{R}^{b \times m \times h}, a \in \{l,r\}$
\end{algorithmic}
\end{algorithm}

%% file: Figures_supp/user_study.tex
\begin{figure*}[t]
    \centering
    \vspace{-15pt}
    \includegraphics[width=\textwidth]{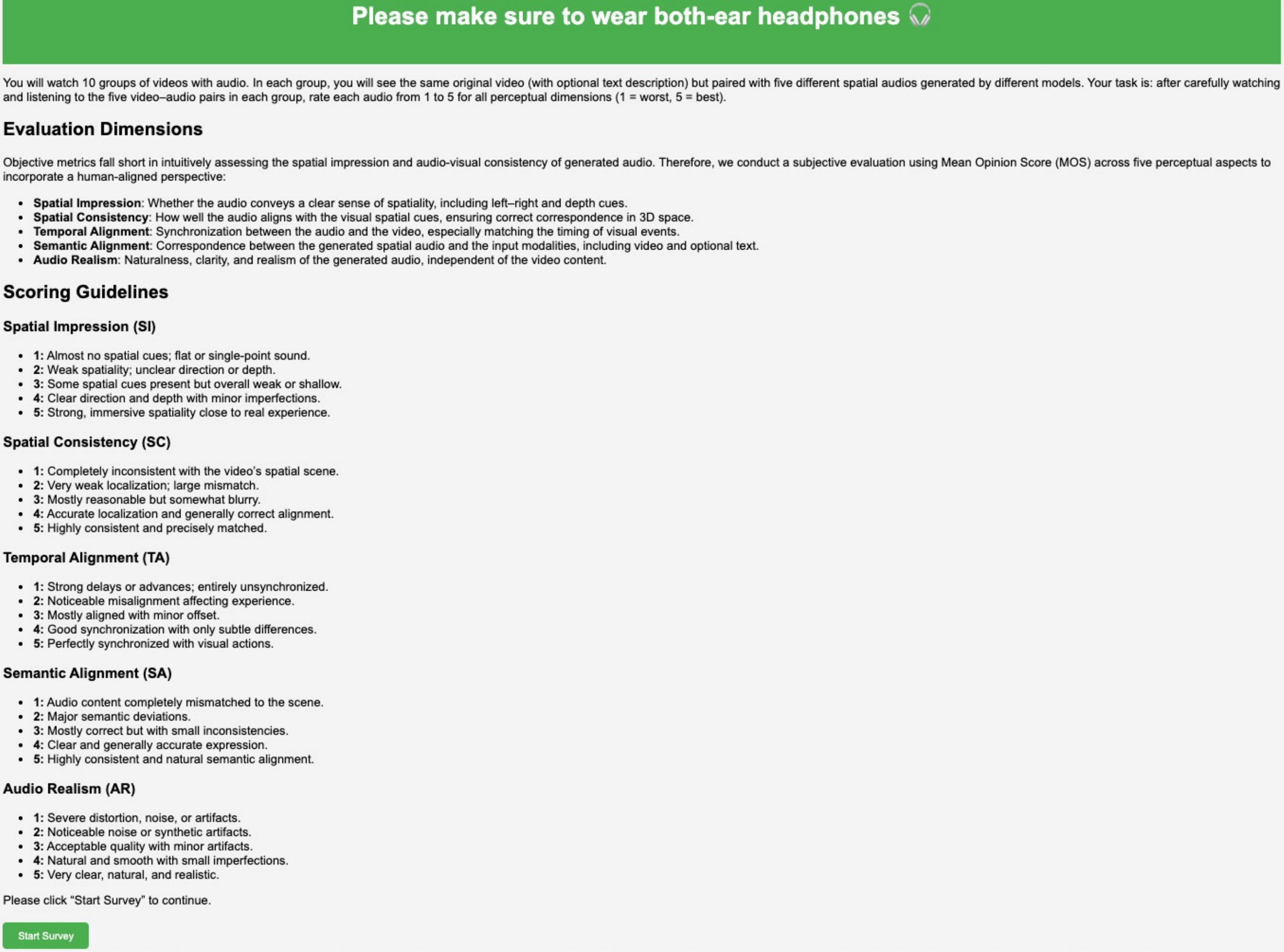}
    \vspace{-20pt}
    \caption{
    \textbf{Subjective Evaluation Criteria.}
    Screenshot of the questionnaire webpage shown to participants, presenting the five perceptual dimensions and their specific scoring guidelines used for the subjective evaluation of spatial audio generation.
    }
    \vspace{-5pt}
    \label{fig:userstudy_a}
\end{figure*}

\begin{figure*}[t]
    \centering
    \includegraphics[width=\textwidth]{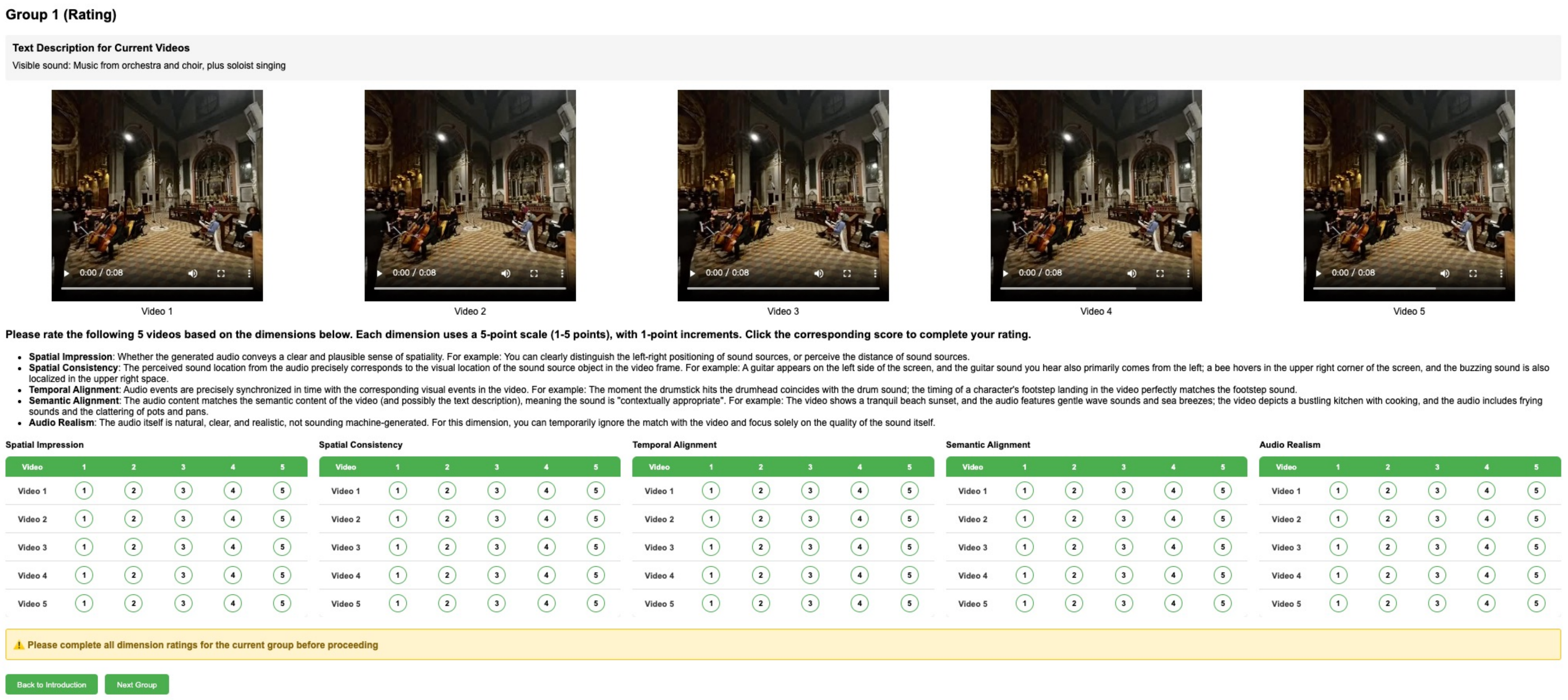}
    \vspace{-20pt}
    \caption{
    \textbf{User Study Rating Interface.}
    Screenshot of the questionnaire webpage used to collect participants’ MOS ratings across the five perceptual dimensions for each video–audio group.
    }
    \vspace{-20pt}
    \label{fig:userstudy_b}
\end{figure*}

%% file: Tables_supp/ablation_module.tex
\begin{table*}[t]
  \centering
  \caption{\textbf{Ablation Study on Key Model Components.} We conduct an ablation study to evaluate the contributions of key model components: \textit{Dual, Dual-Branch Audio Generation} (Sec. 4.2) and \textit{Spt, Conditional Spacetime Module} (Sec. 4.3). \textit{Pretrained} refers to our pretrained MMAudio~\cite{MMAudio} model, spatialized by duplicating the mono output into both channels for evaluation.}
  \vspace{-7pt}
  \label{tab:ablation_module}
  \begin{tabular}{lccccccccc}
    \toprule
    \multirow{2}{*}{\textbf{Model}} &
    \multicolumn{6}{c}{\textbf{Audio Distribution Matching}} &
    \multicolumn{2}{c}{\textbf{Video-Audio Alignment}} \\
    \cmidrule(lr){2-7} \cmidrule(lr){8-9}
    &
    \textbf{$\text{FD}_{\text{VGG}}^{\text{mix}}$$\downarrow$} &
    \textbf{$\text{FD}_{\text{VGG}}^{\text{avg}}$$\downarrow$} &
    \textbf{$\text{FD}_{\text{PANN}}^{\text{mix}}$$\downarrow$} &
    \textbf{$\text{FD}_{\text{PANN}}^{\text{avg}}$$\downarrow$} &
    \textbf{$\text{KL}_{\text{PANN}}^{\text{mix}}$$\downarrow$} &
    \textbf{$\text{KL}_{\text{PANN}}^{\text{avg}}$$\downarrow$} &
    \textbf{DeSync$\downarrow$} &
    \textbf{IB-Score$\uparrow$} \\
    \midrule
    Pretrained & 5.675 & 4.482 & 18.341 & 16.711 & 1.845 & 1.837 & 0.793 & 0.285 \\
    w/ Dual only & 2.908 & 2.803 & 15.151 & 13.008 & 1.646 & \textbf{1.572} & \textbf{0.766} & 0.289 \\
    \rowcolor{lightblue} Dual+Spt & \textbf{2.516} & \textbf{2.479} & \textbf{13.917} & \textbf{12.684} & \textbf{1.638} & 1.573 & 0.788 & \textbf{0.299} \\
    \bottomrule
  \end{tabular}
\end{table*}

%% file: Tables_supp/ablation_dataset.tex
\begin{table*}[t]
  \centering
  \vspace{5pt}
  \caption{\textbf{Ablation Study on Training Dataset.} We conduct an ablation study to evaluate the contributions of our dataset. We train the model both without and with using our \DatasetName{} dataset, and evaluate it on our test set. The results show that without our dataset, the model fails to generate high-quality binaural audio on open-domain data.}
  \vspace{-7pt}
  \label{tab:ablation_dataset}
  \begin{tabular}{lccccccccc}
    \toprule
    \multirow{2}{*}{\textbf{Model}} &
    \multicolumn{6}{c}{\textbf{Audio Distribution Matching}} &
    \multicolumn{2}{c}{\textbf{Video-Audio Alignment}} \\
    \cmidrule(lr){2-7} \cmidrule(lr){8-9}
    &
    \textbf{$\text{FD}_{\text{VGG}}^{\text{mix}}$$\downarrow$} &
    \textbf{$\text{FD}_{\text{VGG}}^{\text{avg}}$$\downarrow$} &
    \textbf{$\text{FD}_{\text{PANN}}^{\text{mix}}$$\downarrow$} &
    \textbf{$\text{FD}_{\text{PANN}}^{\text{avg}}$$\downarrow$} &
    \textbf{$\text{KL}_{\text{PANN}}^{\text{mix}}$$\downarrow$} &
    \textbf{$\text{KL}_{\text{PANN}}^{\text{avg}}$$\downarrow$} &
    \textbf{DeSync$\downarrow$} &
    \textbf{IB-Score$\uparrow$} \\
    \midrule
    w/o BiAudio & 10.966 & 12.677 & 43.711 & 45.175 & 4.203 & 4.280 & 0.916 & 0.102 \\
    \rowcolor{lightblue} w/ BiAudio & \textbf{2.516} & \textbf{2.479} & \textbf{13.917} & \textbf{12.684} & \textbf{1.638} & \textbf{1.573} & \textbf{0.788} & \textbf{0.299} \\
    \bottomrule
  \end{tabular}
\end{table*}

%% file: main.bib
@String(CVPR= {IEEE Conf. Comput. Vis. Pattern Recog.})

@String(ECCV= {Eur. Conf. Comput. Vis.})

@String(BMVC= {Brit. Mach. Vis. Conf.})

@String(ICASSP=	{ICASSP})

@String(ICLR = {Int. Conf. Learn. Represent.})

@String(AAAI = {AAAI})

@String(VR   = {Vis. Res.})

@String(CVPR  = {CVPR})

@String(ECCV  = {ECCV})

@String(BMVC  =	{BMVC})

@String(ICLR  = {ICLR})

@article{OmniAudio,
  author       = {Huadai Liu and
                  Tianyi Luo and
                  Qikai Jiang and
                  Kaicheng Luo and
                  Peiwen Sun and
                  Jialei Wang and
                  Rongjie Huang and
                  Qian Chen and
                  Wen Wang and
                  Xiangtai Li and
                  Shiliang Zhang and
                  Zhijie Yan and
                  Zhou Zhao and
                  Wei Xue},
  title        = {OmniAudio: Generating Spatial Audio from 360-Degree Video},
  journal      = {CoRR},
  volume       = {abs/2504.14906},
  year         = {2025}
}

@article{Qwen2.5-Omni,
  title={Qwen2.5-Omni Technical Report},
  author={Jin Xu and Zhifang Guo and Jinzheng He and Hangrui Hu and Ting He and Shuai Bai and Keqin Chen and Jialin Wang and Yang Fan and Kai Dang and Bin Zhang and Xiong Wang and Yunfei Chu and Junyang Lin},
  journal={arXiv preprint arXiv:2503.20215},
  year={2025}
}

@article{qwen3,
    title={Qwen3 Technical Report}, 
    author={An Yang and Anfeng Li and Baosong Yang and Beichen Zhang and Binyuan Hui and Bo Zheng and Bowen Yu and Chang Gao and Chengen Huang and Chenxu Lv and Chujie Zheng and Dayiheng Liu and Fan Zhou and Fei Huang and Feng Hu and Hao Ge and Haoran Wei and Huan Lin and Jialong Tang and Jian Yang and Jianhong Tu and Jianwei Zhang and Jianxin Yang and Jiaxi Yang and Jing Zhou and Jingren Zhou and Junyang Lin and Kai Dang and Keqin Bao and Kexin Yang and Le Yu and Lianghao Deng and Mei Li and Mingfeng Xue and Mingze Li and Pei Zhang and Peng Wang and Qin Zhu and Rui Men and Ruize Gao and Shixuan Liu and Shuang Luo and Tianhao Li and Tianyi Tang and Wenbiao Yin and Xingzhang Ren and Xinyu Wang and Xinyu Zhang and Xuancheng Ren and Yang Fan and Yang Su and Yichang Zhang and Yinger Zhang and Yu Wan and Yuqiong Liu and Zekun Wang and Zeyu Cui and Zhenru Zhang and Zhipeng Zhou and Zihan Qiu},
    journal = {arXiv preprint arXiv:2505.09388},
    year={2025}
}

@inproceedings{ViSAGe,
  author       = {Jaeyeon Kim and
                  Heeseung Yun and
                  Gunhee Kim},
  title        = {ViSAGe: Video-to-Spatial Audio Generation},
  booktitle    = {{ICLR}},
  publisher    = {OpenReview.net},
  year         = {2025}
}

@inproceedings{YT-360,
  author       = {Pedro Morgado and
                  Nuno Vasconcelos and
                  Timothy R. Langlois and
                  Oliver Wang},
  title        = {Self-Supervised Generation of Spatial Audio for 360{\textdegree} Video},
  booktitle    = {NeurIPS},
  pages        = {360--370},
  year         = {2018}
}

@inproceedings{OAP,
  author       = {Arun Balajee Vasudevan and
                  Dengxin Dai and
                  Luc Van Gool},
  title        = {Semantic Object Prediction and Spatial Sound Super-Resolution with
                  Binaural Sounds},
  booktitle    = {{ECCV} {(4)}},
  series       = {Lecture Notes in Computer Science},
  volume       = {12349},
  pages        = {638--655},
  publisher    = {Springer},
  year         = {2020}
}

@InProceedings{MUSIC1,
    author = {Zhao, Hang and Gan, Chuang and Rouditchenko, Andrew and Vondrick, Carl and McDermott, Josh and Torralba, Antonio},
    title = {The Sound of Pixels},
    booktitle = {The European Conference on Computer Vision (ECCV)},
    month = {September},
    year = {2018}
}

@inproceedings{MUSIC2,
  title={The sound of motions},
  author={Zhao, Hang and Gan, Chuang and Ma, Wei-Chiu and Torralba, Antonio},
  booktitle={Proceedings of the IEEE International Conference on Computer Vision},
  pages={1735--1744},
  year={2019}
}

@inproceedings{Fair-Play,
  title={2.5D Visual Sound},
  author={Gao, Ruohan and Grauman, Kristen},
  booktitle={CVPR},
  year={2019}
}

@article{SimBinaural,
  author       = {Rishabh Garg and
                  Ruohan Gao and
                  Kristen Grauman},
  title        = {Visually-Guided Audio Spatialization in Video with Geometry-Aware
                  Multi-task Learning},
  journal      = {Int. J. Comput. Vis.},
  volume       = {131},
  number       = {10},
  pages        = {2723--2737},
  year         = {2023}
}

@inproceedings{Diffusion,
  author       = {Robin Rombach and
                  Andreas Blattmann and
                  Dominik Lorenz and
                  Patrick Esser and
                  Bj{\"{o}}rn Ommer},
  title        = {High-Resolution Image Synthesis with Latent Diffusion Models},
  booktitle    = {{CVPR}},
  pages        = {10674--10685},
  publisher    = {{IEEE}},
  year         = {2022}
}

@inproceedings{Flow-Matching,
  author       = {Yaron Lipman and
                  Ricky T. Q. Chen and
                  Heli Ben{-}Hamu and
                  Maximilian Nickel and
                  Matthew Le},
  title        = {Flow Matching for Generative Modeling},
  booktitle    = {{ICLR}},
  publisher    = {OpenReview.net},
  year         = {2023}
}

@inproceedings{SpecVQGAN,
  author       = {Vladimir Iashin and
                  Esa Rahtu},
  title        = {Taming Visually Guided Sound Generation},
  booktitle    = {{BMVC}},
  pages        = {2},
  publisher    = {{BMVA} Press},
  year         = {2021}
}

@inproceedings{Foleygen,
  author       = {Xinhao Mei and
                  Varun Nagaraja and
                  Ga{\"{e}}l Le Lan and
                  Zhaoheng Ni and
                  Ernie Chang and
                  Yangyang Shi and
                  Vikas Chandra},
  title        = {Foleygen: Visually-Guided Audio Generation},
  booktitle    = {{MLSP}},
  pages        = {1--6},
  publisher    = {{IEEE}},
  year         = {2024}
}

@inproceedings{V-AURA,
  author       = {Ilpo Viertola and
                  Vladimir Iashin and
                  Esa Rahtu},
  title        = {Temporally Aligned Audio for Video with Autoregression},
  booktitle    = {{ICASSP}},
  pages        = {1--5},
  publisher    = {{IEEE}},
  year         = {2025}
}

@inproceedings{MaskVAT,
  author       = {Santiago Pascual and
                  Chunghsin Yeh and
                  Ioannis Tsiamas and
                  Joan Serr{\`{a}}},
  title        = {Masked Generative Video-to-Audio Transformers with Enhanced Synchronicity},
  booktitle    = {{ECCV} {(87)}},
  series       = {Lecture Notes in Computer Science},
  volume       = {15145},
  pages        = {247--264},
  publisher    = {Springer},
  year         = {2024}
}

@inproceedings{LoVA,
  author       = {Xin Cheng and
                  Xihua Wang and
                  Yihan Wu and
                  Yuyue Wang and
                  Ruihua Song},
  title        = {LoVA: Long-form Video-to-Audio Generation},
  booktitle    = {{ICASSP}},
  pages        = {1--5},
  publisher    = {{IEEE}},
  year         = {2025}
}

@article{VTA-LDM,
  author       = {Manjie Xu and
                  Chenxing Li and
                  Yong Ren and
                  Rilin Chen and
                  Yu Gu and
                  Wei Liang and
                  Dong Yu},
  title        = {Video-to-Audio Generation with Hidden Alignment},
  journal      = {CoRR},
  volume       = {abs/2407.07464},
  year         = {2024}
}

@inproceedings{Seeing-and-Hearing,
  author       = {Yazhou Xing and
                  Yingqing He and
                  Zeyue Tian and
                  Xintao Wang and
                  Qifeng Chen},
  title        = {Seeing and Hearing: Open-domain Visual-Audio Generation with Diffusion
                  Latent Aligners},
  booktitle    = {{CVPR}},
  pages        = {7151--7161},
  publisher    = {{IEEE}},
  year         = {2024}
}

@inproceedings{Frieren,
  author       = {Yongqi Wang and
                  Wenxiang Guo and
                  Rongjie Huang and
                  Jiawei Huang and
                  Zehan Wang and
                  Fuming You and
                  Ruiqi Li and
                  Zhou Zhao},
  title        = {Frieren: Efficient Video-to-Audio Generation Network with Rectified
                  Flow Matching},
  booktitle    = {NeurIPS},
  year         = {2024}
}

@article{Kling-Foley,
  author       = {Jun Wang and
                  Xijuan Zeng and
                  Chunyu Qiang and
                  Ruilong Chen and
                  Shiyao Wang and
                  Le Wang and
                  Wangjing Zhou and
                  Pengfei Cai and
                  Jiahui Zhao and
                  Nan Li and
                  Zihan Li and
                  Yuzhe Liang and
                  Xiaopeng Wang and
                  Haorui Zheng and
                  Ming Wen and
                  Kang Yin and
                  Yiran Wang and
                  Nan Li and
                  Feng Deng and
                  Liang Dong and
                  Chen Zhang and
                  Di Zhang and
                  Kun Gai},
  title        = {Kling-Foley: Multimodal Diffusion Transformer for High-Quality Video-to-Audio
                  Generation},
  journal      = {CoRR},
  volume       = {abs/2506.19774},
  year         = {2025}
}

@inproceedings{Diff-Foley,
  author       = {Simian Luo and
                  Chuanhao Yan and
                  Chenxu Hu and
                  Hang Zhao},
  title        = {Diff-Foley: Synchronized Video-to-Audio Synthesis with Latent Diffusion
                  Models},
  booktitle    = {NeurIPS},
  year         = {2023}
}

@article{MovieGen,
  author       = {Adam Polyak and
                  Amit Zohar and
                  Andrew Brown and
                  Andros Tjandra and
                  Animesh Sinha and
                  Ann Lee and
                  Apoorv Vyas and
                  Bowen Shi and
                  Chih{-}Yao Ma and
                  Ching{-}Yao Chuang and
                  David Yan and
                  Dhruv Choudhary and
                  Dingkang Wang and
                  Geet Sethi and
                  Guan Pang and
                  Haoyu Ma and
                  Ishan Misra and
                  Ji Hou and
                  Jialiang Wang and
                  Kiran Jagadeesh and
                  Kunpeng Li and
                  Luxin Zhang and
                  Mannat Singh and
                  Mary Williamson and
                  Matt Le and
                  Matthew Yu and
                  Mitesh Kumar Singh and
                  Peizhao Zhang and
                  Peter Vajda and
                  Quentin Duval and
                  Rohit Girdhar and
                  Roshan Sumbaly and
                  Sai Saketh Rambhatla and
                  Sam S. Tsai and
                  Samaneh Azadi and
                  Samyak Datta and
                  Sanyuan Chen and
                  Sean Bell and
                  Sharadh Ramaswamy and
                  Shelly Sheynin and
                  Siddharth Bhattacharya and
                  Simran Motwani and
                  Tao Xu and
                  Tianhe Li and
                  Tingbo Hou and
                  Wei{-}Ning Hsu and
                  Xi Yin and
                  Xiaoliang Dai and
                  Yaniv Taigman and
                  Yaqiao Luo and
                  Yen{-}Cheng Liu and
                  Yi{-}Chiao Wu and
                  Yue Zhao and
                  Yuval Kirstain and
                  Zecheng He and
                  Zijian He and
                  Albert Pumarola and
                  Ali K. Thabet and
                  Artsiom Sanakoyeu and
                  Arun Mallya and
                  Baishan Guo and
                  Boris Araya and
                  Breena Kerr and
                  Carleigh Wood and
                  Ce Liu and
                  Cen Peng and
                  Dmitry Vengertsev and
                  Edgar Sch{\"{o}}nfeld and
                  Elliot Blanchard and
                  Felix Juefei{-}Xu and
                  Fraylie Nord and
                  Jeff Liang and
                  John Hoffman and
                  Jonas Kohler and
                  Kaolin Fire and
                  Karthik Sivakumar and
                  Lawrence Chen and
                  Licheng Yu and
                  Luya Gao and
                  Markos Georgopoulos and
                  Rashel Moritz and
                  Sara K. Sampson and
                  Shikai Li and
                  Simone Parmeggiani and
                  Steve Fine and
                  Tara Fowler and
                  Vladan Petrovic and
                  Yuming Du},
  title        = {Movie Gen: {A} Cast of Media Foundation Models},
  journal      = {CoRR},
  volume       = {abs/2410.13720},
  year         = {2024}
}

@article{See-Hear-Read,
  author       = {Yusuf Aytar and
                  Carl Vondrick and
                  Antonio Torralba},
  title        = {See, Hear, and Read: Deep Aligned Representations},
  journal      = {CoRR},
  volume       = {abs/1706.00932},
  year         = {2017}
}

@article{FoleyCrafter,
  author       = {Yiming Zhang and
                  Yicheng Gu and
                  Yanhong Zeng and
                  Zhening Xing and
                  Yuancheng Wang and
                  Zhizheng Wu and
                  Kai Chen},
  title        = {FoleyCrafter: Bring Silent Videos to Life with Lifelike and Synchronized
                  Sounds},
  journal      = {CoRR},
  volume       = {abs/2407.01494},
  year         = {2024}
}

@inproceedings{CODI2,
  author       = {Zineng Tang and
                  Ziyi Yang and
                  Mahmoud Khademi and
                  Yang Liu and
                  Chenguang Zhu and
                  Mohit Bansal},
  title        = {CoDi-2: In-Context, Interleaved, and Interactive Any-to-Any Generation},
  booktitle    = {{CVPR}},
  pages        = {27415--27424},
  publisher    = {{IEEE}},
  year         = {2024}
}

@inproceedings{MMAudio,
  author       = {Ho Kei Cheng and
                  Masato Ishii and
                  Akio Hayakawa and
                  Takashi Shibuya and
                  Alexander G. Schwing and
                  Yuki Mitsufuji},
  title        = {MMAudio: Taming Multimodal Joint Training for High-Quality Video-to-Audio
                  Synthesis},
  booktitle    = {{CVPR}},
  pages        = {28901--28911},
  publisher    = {Computer Vision Foundation / {IEEE}},
  year         = {2025}
}

@article{ThinkSound,
  author       = {Huadai Liu and
                  Jialei Wang and
                  Kaicheng Luo and
                  Wen Wang and
                  Qian Chen and
                  Zhou Zhao and
                  Wei Xue},
  title        = {ThinkSound: Chain-of-Thought Reasoning in Multimodal Large Language
                  Models for Audio Generation and Editing},
  journal      = {CoRR},
  volume       = {abs/2506.21448},
  year         = {2025}
}

@inproceedings{MultiFoley,
  author       = {Ziyang Chen and
                  Prem Seetharaman and
                  Bryan C. Russell and
                  Oriol Nieto and
                  David Bourgin and
                  Andrew Owens and
                  Justin Salamon},
  title        = {Video-Guided Foley Sound Generation with Multimodal Controls},
  booktitle    = {{CVPR}},
  pages        = {18770--18781},
  publisher    = {Computer Vision Foundation / {IEEE}},
  year         = {2025}
}

@inproceedings{TA-V2A,
  author       = {Yuhuan You and
                  Xihong Wu and
                  Tianshu Qu},
  title        = {{TA-V2A:} Textually Assisted Video-to-Audio Generation},
  booktitle    = {{ICASSP}},
  pages        = {1--5},
  publisher    = {{IEEE}},
  year         = {2025}
}

@inproceedings{VQGAN,
  author       = {Patrick Esser and
                  Robin Rombach and
                  Bj{\"{o}}rn Ommer},
  title        = {Taming Transformers for High-Resolution Image Synthesis},
  booktitle    = {{CVPR}},
  pages        = {12873--12883},
  publisher    = {Computer Vision Foundation / {IEEE}},
  year         = {2021}
}

@inproceedings{RFM,
  author       = {Xingchao Liu and
                  Chengyue Gong and
                  Qiang Liu},
  title        = {Flow Straight and Fast: Learning to Generate and Transfer Data with
                  Rectified Flow},
  booktitle    = {{ICLR}},
  publisher    = {OpenReview.net},
  year         = {2023}
}

@article{Ccstereo,
  title={Ccstereo: Audio-visual contextual and contrastive learning for binaural audio generation},
  author={Chen, Yuanhong and Shimada, Kazuki and Simon, Christian and Ikemiya, Yukara and Shibuya, Takashi and Mitsufuji, Yuki},
  journal={arXiv preprint arXiv:2501.02786},
  year={2025}
}

@inproceedings{See-2-sound,
  title={See-2-sound: Zero-shot spatial environment-to-spatial sound},
  author={Dagli, Rishit and Prakash, Shivesh and Wu, Robert and Khosravani, Houman},
  booktitle={Proceedings of the Special Interest Group on Computer Graphics and Interactive Techniques Conference Posters},
  pages={1--2},
  year={2025}
}

@article{FoleySpace,
  title={FoleySpace: Vision-Aligned Binaural Spatial Audio Generation},
  author={Zhao, Lei and Chen, Rujin and Zhang, Chi and Zhang, Xiao-Lei and Li, Xuelong},
  journal={arXiv preprint arXiv:2508.12918},
  year={2025}
}

@article{Sonic4D,
  title={Sonic4D: Spatial Audio Generation for Immersive 4D Scene Exploration},
  author={Xie, Siyi and Zhu, Hanxin and He, Tianyu and Li, Xin and Chen, Zhibo},
  journal={arXiv preprint arXiv:2506.15759},
  year={2025}
}

@article{Geometry,
  title={Geometry-aware multi-task learning for binaural audio generation from video},
  author={Garg, Rishabh and Gao, Ruohan and Grauman, Kristen},
  journal={arXiv preprint arXiv:2111.10882},
  year={2021}
}

@article{SpatialAudioGen,
  title={Self-supervised generation of spatial audio for 360 video},
  author={Morgado, Pedro and Nvasconcelos, Nuno and Langlois, Timothy and Wang, Oliver},
  journal={Advances in neural information processing systems},
  volume={31},
  year={2018}
}

@inproceedings{Mono2Binaural,
  title={2.5 d visual sound},
  author={Gao, Ruohan and Grauman, Kristen},
  booktitle={Proceedings of the IEEE/CVF Conference on Computer Vision and Pattern Recognition},
  pages={324--333},
  year={2019}
}

@inproceedings{Sep-stereo,
  title={Sep-stereo: Visually guided stereophonic audio generation by associating source separation},
  author={Zhou, Hang and Xu, Xudong and Lin, Dahua and Wang, Xiaogang and Liu, Ziwei},
  booktitle={European Conference on Computer Vision},
  pages={52--69},
  year={2020},
  organization={Springer}
}

@inproceedings{PseudoBinaural,
  title={Visually informed binaural audio generation without binaural audios},
  author={Xu, Xudong and Zhou, Hang and Liu, Ziwei and Dai, Bo and Wang, Xiaogang and Lin, Dahua},
  booktitle={Proceedings of the IEEE/CVF Conference on Computer Vision and Pattern Recognition},
  pages={15485--15494},
  year={2021}
}

@inproceedings{Unet,
  title={U-net: Convolutional networks for biomedical image segmentation},
  author={Ronneberger, Olaf and Fischer, Philipp and Brox, Thomas},
  booktitle={International Conference on Medical image computing and computer-assisted intervention},
  pages={234--241},
  year={2015},
  organization={Springer}
}

@article{Augmented,
  title={Improving Binaural Audio Techniques for Augmented Reality},
  author={Alonso-Mart{\i}nez, Juan Isaac Engel},
  year={2021}
}

@inproceedings{virtual,
  title={Binaural sound reduces reaction time in a virtual reality search task},
  author={Hoeg, Emil R and Gerry, Lynda J and Thomsen, Lui and Nilsson, Niels C and Serafin, Stefania},
  booktitle={2017 IEEE 3rd VR workshop on sonic interactions for virtual environments (SIVE)},
  pages={1--4},
  year={2017},
  organization={IEEE}
}

@book{Ambisonics,
  title={Ambisonics: A practical 3D audio theory for recording, studio production, sound reinforcement, and virtual reality},
  author={Zotter, Franz and Frank, Matthias},
  year={2019},
  publisher={Springer Nature}
}

@article{AudioX,
  title={AudioX: Diffusion Transformer for Anything-to-Audio Generation},
  author={Tian, Zeyue and Jin, Yizhu and Liu, Zhaoyang and Yuan, Ruibin and Tan, Xu and Chen, Qifeng and Xue, Wei and Guo, Yike},
  journal={arXiv preprint arXiv:2503.10522},
  year={2025}
}

@article{FM2,
  author       = {Alexander Tong and
                  Kilian Fatras and
                  Nikolay Malkin and
                  Guillaume Huguet and
                  Yanlei Zhang and
                  Jarrid Rector{-}Brooks and
                  Guy Wolf and
                  Yoshua Bengio},
  title        = {Improving and generalizing flow-based generative models with minibatch
                  optimal transport},
  journal      = {Trans. Mach. Learn. Res.},
  volume       = {2024},
  year         = {2024}
}

@article{OT,
  title={A convexity principle for interacting gases},
  author={McCann, Robert J},
  journal={Advances in mathematics},
  volume={128},
  number={1},
  pages={153--179},
  year={1997},
  publisher={Elsevier}
}

@inproceedings{LAD,
  title={Fast timing-conditioned latent audio diffusion},
  author={Evans, Zach and Carr, CJ and Taylor, Josiah and Hawley, Scott H and Pons, Jordi},
  booktitle={Forty-first International Conference on Machine Learning},
  year={2024}
}

@inproceedings{mono-vae,
  author       = {Diederik P. Kingma and
                  Max Welling},
  title        = {Auto-Encoding Variational Bayes},
  booktitle    = {{ICLR}},
  year         = {2014}
}

@inproceedings{Vocoder,
  author       = {Sang{-}gil Lee and
                  Wei Ping and
                  Boris Ginsburg and
                  Bryan Catanzaro and
                  Sungroh Yoon},
  title        = {BigVGAN: {A} Universal Neural Vocoder with Large-Scale Training},
  booktitle    = {{ICLR}},
  publisher    = {OpenReview.net},
  year         = {2023}
}

@inproceedings{clip,
  author       = {Alec Radford and
                  Jong Wook Kim and
                  Chris Hallacy and
                  Aditya Ramesh and
                  Gabriel Goh and
                  Sandhini Agarwal and
                  Girish Sastry and
                  Amanda Askell and
                  Pamela Mishkin and
                  Jack Clark and
                  Gretchen Krueger and
                  Ilya Sutskever},
  title        = {Learning Transferable Visual Models From Natural Language Supervision},
  booktitle    = {{ICML}},
  series       = {Proceedings of Machine Learning Research},
  volume       = {139},
  pages        = {8748--8763},
  publisher    = {{PMLR}},
  year         = {2021}
}

@inproceedings{Synchformer,
  author       = {Vladimir Iashin and
                  Weidi Xie and
                  Esa Rahtu and
                  Andrew Zisserman},
  title        = {Synchformer: Efficient Synchronization From Sparse Cues},
  booktitle    = {{ICASSP}},
  pages        = {5325--5329},
  publisher    = {{IEEE}},
  year         = {2024}
}

@article{pe,
  author       = {Daniel Bolya and
                  Po{-}Yao Huang and
                  Peize Sun and
                  Jang Hyun Cho and
                  Andrea Madotto and
                  Chen Wei and
                  Tengyu Ma and
                  Jiale Zhi and
                  Jathushan Rajasegaran and
                  Hanoona Rasheed and
                  Junke Wang and
                  Marco Monteiro and
                  Hu Xu and
                  Shiyu Dong and
                  Nikhila Ravi and
                  Daniel Li and
                  Piotr Doll{\'{a}}r and
                  Christoph Feichtenhofer},
  title        = {Perception Encoder: The best visual embeddings are not at the output
                  of the network},
  journal      = {CoRR},
  volume       = {abs/2504.13181},
  year         = {2025}
}

@misc{FLUX,
      title={FLUX.1 Kontext: Flow Matching for In-Context Image Generation and Editing in Latent Space},
      author={Black Forest Labs and Stephen Batifol and Andreas Blattmann and Frederic Boesel and Saksham Consul and Cyril Diagne and Tim Dockhorn and Jack English and Zion English and Patrick Esser and Sumith Kulal and Kyle Lacey and Yam Levi and Cheng Li and Dominik Lorenz and Jonas Müller and Dustin Podell and Robin Rombach and Harry Saini and Axel Sauer and Luke Smith},
      year={2025},
      eprint={2506.15742},
      archivePrefix={arXiv},
      primaryClass={cs.GR},
      url={https://arxiv.org/abs/2506.15742},
}

@article{PANNs,
  title={PANNs: Large-Scale Pretrained Audio Neural Networks for Audio Pattern Recognition},
  author={Qiuqiang Kong and Yin Cao and Turab Iqbal and Yuxuan Wang and Wenwu Wang and Mark D. Plumbley},
  journal={IEEE/ACM Transactions on Audio, Speech, and Language Processing},
  year={2019},
  volume={28},
  pages={2880-2894},
  url={https://api.semanticscholar.org/CorpusID:209444382}
}

@inproceedings{AudioSet,
  author       = {Jort F. Gemmeke and
                  Daniel P. W. Ellis and
                  Dylan Freedman and
                  Aren Jansen and
                  Wade Lawrence and
                  R. Channing Moore and
                  Manoj Plakal and
                  Marvin Ritter},
  title        = {Audio Set: An ontology and human-labeled dataset for audio events},
  booktitle    = {{ICASSP}},
  pages        = {776--780},
  publisher    = {{IEEE}},
  year         = {2017}
}

@inproceedings{ImageBind,
  author       = {Rohit Girdhar and
                  Alaaeldin El{-}Nouby and
                  Zhuang Liu and
                  Mannat Singh and
                  Kalyan Vasudev Alwala and
                  Armand Joulin and
                  Ishan Misra},
  title        = {ImageBind One Embedding Space to Bind Them All},
  booktitle    = {{CVPR}},
  pages        = {15180--15190},
  publisher    = {{IEEE}},
  year         = {2023}
}

@InProceedings{adaLN,
  title={FiLM: Visual Reasoning with a General Conditioning Layer},
  author={Ethan Perez and Florian Strub and Harm de Vries and Vincent Dumoulin and Aaron C. Courville},
  booktitle={AAAI},
  year={2018}
}

@software{pyequilib2021github,
  author = {Haruya Ishikawa},
  title = {PyEquilib: Processing Equirectangular Images with Python},
  url = {http://github.com/haruishi43/equilib},
  version = {0.5.0},
  year = {2021},
}

@software{Omnitone,
  author = {GoogleChrome},
  title = {Omnitone: Spatial Audio Rendering on the Web},
  url = {https://github.com/GoogleChrome/omnitone},
  year = {2016},
}
